\documentclass[journal]{IEEEtran}
\usepackage{ifpdf}
\usepackage{cite}
\usepackage{amsmath}
\usepackage{algorithmic}
\usepackage{array}
\usepackage{epsfig}
\usepackage[switch]{lineno}
\usepackage{float}
\usepackage{graphicx}
\usepackage{color}
\usepackage{amsfonts}
\usepackage{subfigure}
\usepackage{diagbox}
\usepackage{multirow}
\usepackage{setspace}
\begin{document}
    \title{Foreground-background Parallel Compression with Residual Encoding for Surveillance Video}
    
    \author{Lirong Wu, \IEEEmembership{Student Member, IEEE}, Kejie Huang, \IEEEmembership{Senior Member, IEEE}, Haibin Shen, and Lianli Gao
        
        %\thanks{Manuscript received April 19, 2020; revised August 26, 2020.}
        \thanks{Copyright © 2020 IEEE. Personal use of this material is permitted. However, permission to use this material for any other purposes must be obtained from the IEEE by sending an email to pubs-permissions@ieee.org.}
        \thanks{Lirong Wu, Kejie Huang, and Haibin Shen are with the Department of Information Science \& Electronic Engineering, Zhejiang University, Hangzhou 310027, China, e-mail: \{wulirong, huangkejie, shen\_hb\}@zju.edu.cn.}
        \thanks{Lianli Gao is with the School of Computer Science and Engineering, University of Electronic Science and Technology of China, Chengdu, 611731, China, e-mail: lianli.gao@uestc.edu.cn}}
    
    %\markboth{IEEE TRANSACTIONS ON CIRCUITS AND SYSTEMS FOR VIDEO TECHNOLOGY, Vol. 1, No. 1, August 2020}
    
    \maketitle
    %\pagewiselinenumbers
    \begin{abstract}
        The data storage has been one of the bottlenecks in surveillance systems. The conventional video compression schemes such as H.264 and H.265 do not fully utilize the low information density characteristic of the surveillance video, and they attach equal importance to foreground and background when performing compression. In this paper, we propose a novel video compression scheme that compresses the foreground and background of the surveillance video separately. The compression ratio is greatly improved by sharing background information among adjacent frames through an adaptive background updating and interpolation module. Besides, we present two different schemes to compress the foreground and compare their performance in the ablation study to show the importance of temporal information for video compression. In the decoding end, a coarse-to-fine two-stage module is applied to achieve the composition of the foreground and background and the enhancements of frame quality. The experimental results show that our proposed method requires 49.75\% less bpp (bits per pixel) than the conventional algorithm H.265 to achieve the same PSNR (36 dB) on the HEVC dataset.
    \end{abstract}
    
    \begin{IEEEkeywords}
        Surveillance video, background modeling, deep neural network, video coding.
    \end{IEEEkeywords}
    
    \IEEEpeerreviewmaketitle
    
    \section{Introduction}
    \IEEEPARstart{N}{owadays}, video content contributes more than 90\% of network traffic, and this percentage is expected to further increase in the future \cite{networking2016forecast}. Among various videos, surveillance video is expanding exponentially due to the growing concerns for home and city security. With more data being created by video content and analytics, the compression and storage of surveillance videos have become an increasingly hot research topic. In the past decades, mainstream video compression algorithms H.264 \cite{wiegand2003overview} and H.265 \cite{sullivan2012overview} reduce the redundancy of temporal information through handcrafted modules, e.g., block-based motion estimation and Discrete Cosine Transform (DCT). Recently, many DNN-based video compression algorithms are under focused development to achieve higher compression ratio through large-scale end-to-end training and highly non-linear transform \cite{kim2018adversarial,lu2019dvc,wu2018video,yang2018multi}. However, DNN-based algorithms still have room to improve in the field of surveillance video compression, because they do not fully utilize the low information density characteristic in the video, such as unmanned streets and static backgrounds. Besides, there is strong temporal relevance between adjacent frames in the surveillance video and each frame can be divided into two parts: foreground and background. The violent fluctuations of video content often occur in the foreground regions, while the content of the background is constant and static. Therefore, it is not necessary to attach equal importance to the background as to foreground during compression, which may inevitably lead to spatial redundancy.
    
    In this paper, we propose a novel method to greatly improve the compression ratio of surveillance video while keeping the similar video quality. To achieve this, the foreground and background are extracted and compressed separately from surveillance video and the compression bitstreams are indexed to facilitate retrieval. To compress the foreground, we propose a motion-based method with residual encoding, which estimates, compensates and predicts the motion of the foreground objects based on the optical flow information and then encodes the residuals. Besides, the ideas of adaptive background updating and interpolation are applied to background compression to automatically make a trade-off between distortion and compression ratio. Compared with the handcrafted settings, the adaptive scheme overcomes the shortcoming of finding a suitable fixed updating frequency.
    
    As for the decompression of surveillance video, a novel coarse-to-fine, two-stage module is proposed in this paper. The first stage is a composite module that composites the foreground and background into a complete frame. The second stage is a reconstruction network (Rec-Net) that optimizes the composite frames by eliminating boundaries, blurring, and other distortion. The experimental results show that our model outperforms the widely used algorithms, such as H.264 and H.265 when measured by the Multiscale Structural Similarity Index (MS-SSIM) \cite{wang2003multiscale}, Peak Signal to Noise Ratio (PSNR) and subjective perceptual quality. Our contributions are listed in the following:
    
    \begin{itemize}       
        \item We propose a DNN-based method to compress the foreground and background of surveillance video separately, which can be trained in an end-to-end manner.
        \item An adaptive background updating and interpolation algorithm is presented to replace tedious handcrafted settings.
        \item A motion-based method with residual encoding is applied to achieve foreground compression and then the importance of temporal information for video compression is emphasized by the ablation experiment.
        \item The video composition and reconstruction are achieved via a coarse-to-fine, two-stage decoding module.
    \end{itemize}
    
    The rest of the paper is organized as follows: In Section~\ref{sec:2}, some common video compression algorithms and techniques are briefly reviewed. Section~\ref{sec:3} describes the entire architecture of our model and the details of some key components. In Section~\ref{sec:4} and Section~\ref{sec:5}, we present our implementation details and experimental results, respectively. Section~\ref{sec:6} analyzes and summarizes our results, and Section~\ref{sec:7} draws the conclusion.

    \section{Related Work}\label{sec:2}
    \subsection{Conventional Video Compression}
    H.264 and H.265 have been widely used for video compression in recent decades. They have a similar architecture, which contains three types of frames: I-frames, P-frames, and B-frames. Recently, many algorithms specially designed for surveillance videos are under focused development on this basis. For example, Zhang et al. proposed a Background-Modeling Adaptive Prediction (BMAP) method \cite{zhang2013background}. Its basic idea is to adopt different prediction methods based on the classification results of blocks. Unlike this, the Background Difference Coding (BDC) method directly encodes the difference between the input frame and the reconstructed background modeled from the original video \cite{zhang2010efficient}. In addition, Chen et al. proposed a Block-Composed Background Reference (BCBR) method to generate the background reference gradually by block updating instead of picture updating \cite{chen2016block}. The proposed background reference is implemented into H.265, and the experimental results demonstrate a significant improvement in coding efficiency. Besides, a background modeling and referencing scheme for moving cameras-captured surveillance video coding in H.265 was proposed by \cite{wang2018background}. Unlike these, Ma et al. built a library of foreground (vehicles) to search for similar vehicles and used the retrieved results to help encode \cite{ma2019traffic}. However, library-based methods are often not practical because it is difficult to build multiple libraries for all objects that may appear in the surveillance video. The above methods all follow the conventional architecture, which includes components such as motion estimation, motion compensation, etc.  
    
    \subsection{DNN-based Video Compression}
    In recent years, DNN-based video compression is under intensive development. A common approach to adopt DNN in video compression is to improve and optimize a certain component in conventional video compression architecture. The end-to-end Deep Video Compression (DVC) framework \cite{lu2019dvc} proposed by Lu et al. replaced the components of conventional architecture such as motion estimation and motion compensation with DNNs to improve performance. Based on DVC, Lin et al. used multiple reference frames and multiple motion vector fields to generate more accurate predicted frames, resulting in smaller residuals [15].  In contrast to \cite{lu2019dvc} and \cite{lin2020m}, which applied motion estimation and compensation to the whole image, we do not attach equal importance to the foreground and background in the compression process \cite{lin2020m}. Instead, motion estimation and compensation are applied only to the more dramatically changing foreground regions. The relatively static and smooth background regions, on the other hand, are compressed by means of template updates and interpolation, which helps save computation and compression bits. Besides, Chen et al. proposed the concept of PixelMotionCNN (PMCNN) and used a block-based approach to effectively deal with the aforementioned bottleneck of motion compensation trained into a neural network. Instead, we utilize a CNN to extract optical flow as motion information to achieve pixel-level motion compensation \cite{chen2019learning}.
    
    In a very different approach to the problem,  Wu et al. proposed the first an RNN-based Key-Frame Interpolation (KFI) method to generate non-key frames by keyframe interpolation and finally synthesized a complete video \cite{wu2018video}. However, the RNN-based scheme is computationally expensive as several iterations are required and multiple models are needed for different interpolation intervals. Instead, in the interpolation model proposed by Djelouah et al., the required information to generate non-key frames is encoded into a latent representation that directly decodes into motion and blending coefficients, which helps to reduce computation and parameters at the decoding end \cite{djelouah2019neural}. In \cite{Yang2020Learning}, a Hierarchical Learned Video Compression (HLVC) method with three hierarchical quality layers and a recurrent enhancement network was proposed to compress videos according to video quality. Similar to \cite{Yang2020Learning}, the hierarchical compression based on quality is also utilized in our scheme. The main difference is that the idea of hierarchization is only invoked in the background part of our scheme, which divides the background compression into template updates (keyframe compression) and non-key frame interpolation.
    
    \begin{figure*}[ht]
    	\begin{center}
    		\includegraphics[width=\linewidth]{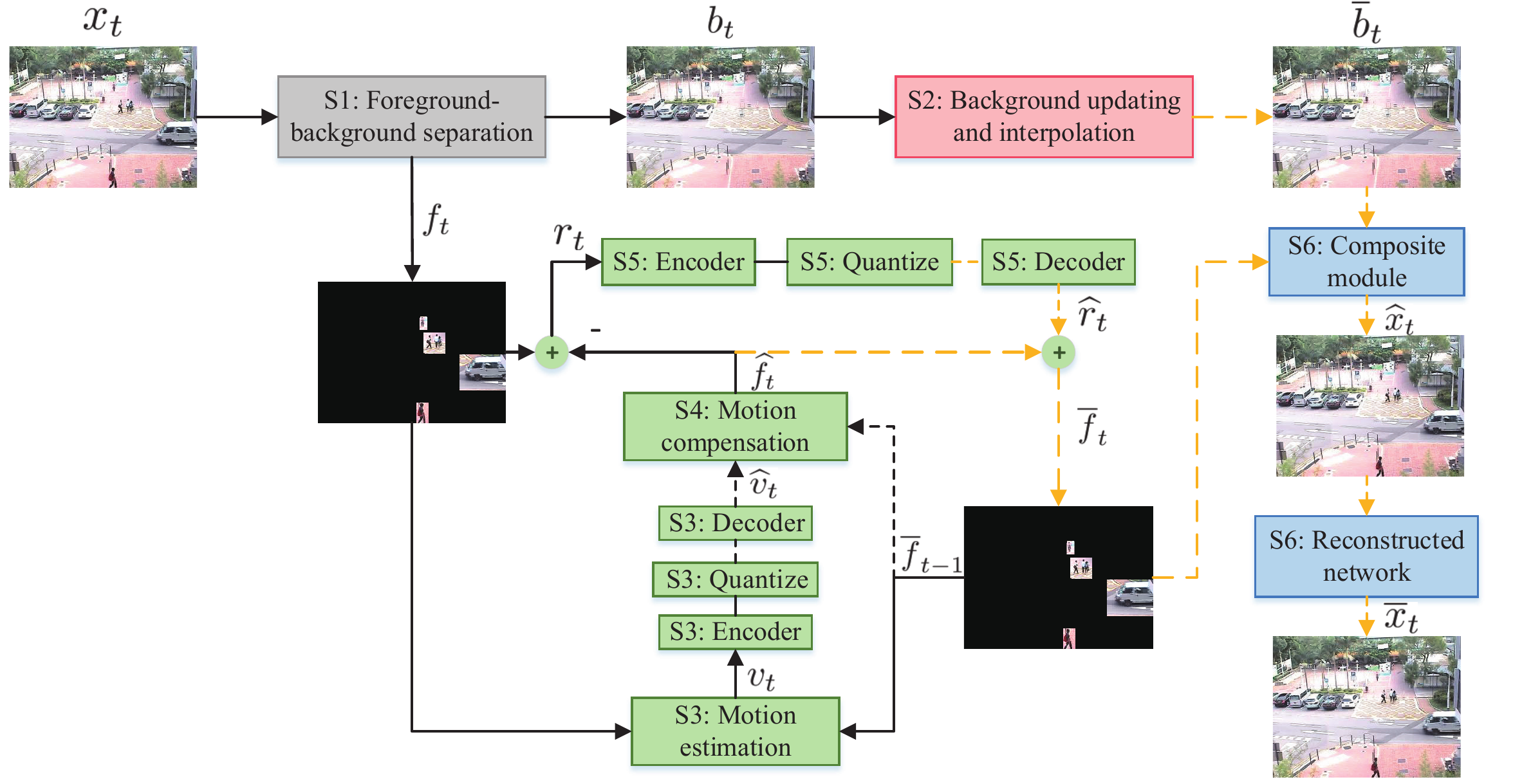}
    	\end{center}
    	\caption{Illustration of our video compression architecture. The {\color{red}red} (S2) and {\color{green}green} (S2-S5)modules are responsible for the compression of background and foreground, respectively. The {\color[rgb]{0.3,0.3,0.3}gray} (S1) and {\color{blue}blue} (S6) modules represent the seperation and composition of foreground and background, respectively. It should be noted that the black lines denote the encoding process, the imaginary lines denote the decoding precess, and the black imaginary lines exist in both process.}
    	\label{fig:1}
    \end{figure*}

    Since Generative Adversarial Network (GAN) is proposed \cite{goodfellow2014generative}, the idea of adversarial training has been applied to the field of compression \cite{santurkar2018generative,lee2018stochastic,agustsson2018generative}. For example, Kim et al. applied GAN to frame prediction and interpolation to achieve video compression \cite{kim2018adversarial}. However, long-term prediction and interpolation can cause severe distortion, so this interpolation method heavily relies on handcrafted settings, such as the selection of keyframes. Afonso et al. proposed ViSTRA, a spatio-temporal resolution adaptation framework for video compression, which dynamically predicted the optimal spatial and temporal resolutions for the input video during encoding and attempted to reconstruct the full resolution video at the decoder. The idea of such an adaptation is also reflected in our work, but the difference is that our resolution is invariant, and our approach is a content-based adaptation, i.e., different compression strategies are applied for different regions of frames \cite{afonso2018video}. Furthermore, Yang et al. proposed a Multi-Frame Quality Enhancement (MFQE) algorithm that used high-quality frames to enhance the quality of low-quality frames based on the fluctuations of compressed video frame quality \cite{yang2018multi}. However, all above DNN-based methods tend to leverage the powerful capabilities of DNNs to find a generic compression framework to handle all scenarios, which disregard the characteristics of the video itself. We believe the compression performance can be greatly improved by using scenario-specific compression methods. Shih et al. reviewed several video compression algorithms specifically designed for sports scenarios, highlighting the broad prospects of scenario-specific compression. In contrast, our method is specifically for surveillance video, which has more complex objects, fluctuations and data volume than sports scenarios \cite{shih2017survey}. Due to the tremendous difference between foreground and background in surveillance videos, they should be compressed parallelly and separately according to their properties. To the best of our knowledge, there is no DNN-based method for foreground-background separated compression.
 
    \subsection{Video Background Modeling}
    The background modeling can separate the foreground and background of the frame and generate an image with background content only. This background image can be used as a long-term reference to encode subsequent frames. A simple method is to treat the long-term average of several frames as the background image. However, this method requires a large number of frames and often fails when the foreground objects remain stationary.
    
    To solve the above problems, there are two types of background modeling methods: non-parameterized \cite{elgammal2000non,liu2003efficient,elgammal2003efficient} and parameterized \cite{stauffer1999adaptive,friedman1997image,klare2009background,kaewtrakulpong2002improved}. For example, Elgammal et al. proposed a nonparametric method to estimate the probability density of a given pixel using kernel density estimation based on historical data \cite{elgammal2000non}. Non-parameterized methods often perform well in highly dynamic scenes, but they are not efficient enough due to the high computational complexity. In real-time applications such as object tracking and video compression, many parameterized methods are applied. For example, the Gaussian Mixture Model (GMM) was proposed, which modeled each pixel of a scene independently through a mixture of four Gaussian distributions \cite{stauffer1999adaptive}. However, these methods often require handcrafted settings of specific parameters based on the application scenario. For this reason, many improved versions have been proposed, such as the improved GMM with adaptive parameters used in this paper \cite{kaewtrakulpong2002improved}.

    \section{Model Architecture}\label{sec:3}
    Fig.~\ref{fig:1} provides a high-level overview of our video compression architecture. The {\color[rgb]{0.3,0.3,0.3}gray} (S1) module performs the separation of foreground and background. The {\color{red}red} (S2) and {\color{green}green} (S3-S5) modules are responsible for the compression of background and foreground, respectively. The {\color{blue}blue} (S6) module represents a coarse-to-fine two-stage module, which achieves the composition of foreground and background and the enhancements of frame quality. It should be noted that the black lines denote the encoding process, the imaginary lines denote the decoding process and the black imaginary lines exist in both processes.
    Let $\mathbb{V}=\{x_{1}, x_{2}, \dots, x_{t-1}, x_{t}, \dots\}$ represents a video sequence, where $x_{t}$ represents the frame at time $t$. $f_{t}$ and $b_{t}$ are the foreground and background separated from $x_{t}$, and $\overline{f}_{t}$ and $\overline{b}_{t}$ are their reconstructed results. $\widehat{x}_{t}$ is the composition of $\overline{f}_{t}$ and $\overline{b}_{t}$, and $\widehat{x}_{t}$ is further post-processed to get $\overline{x}_{t}$. The overall procedure is described in section.~\ref{sec:3.1}, and some important components will be detailed in the remaining sections.
    
    \subsection{Overall Procedure}\label{sec:3.1}
    \noindent\textbf{Step 1 Foreground-background separation.} 
    The adaptive Gaussian mixture model \cite{kaewtrakulpong2002improved} is applied to separate frame $x_{t}$ into background $b_{t}$ and messy foreground points sequentially. Then the foreground points are further processed to obtain foreground $f_{t}$. The specific processing steps are in Sec.~\ref{sec:3.2}.
    
    \noindent\textbf{Step 2 Adaptive background updating and interpolation.} 
    Since the context of the background is generally constant, we can improve the compression ratio by sharing the background information among adjacent frames. This is achieved by an adaptive background updating and interpolation module, where the current background $b_{t}$ is the input of Step2 to produce the reconstructed background $\overline{b}_{t}$. This step will be described in detail in Sec.~\ref{sec:3.3}.
    
    \noindent\textbf{Step 3 Motion estimation.} 
    Instead of conventional block-based motion estimation, we use the optical flow for motion estimation. The corresponding motion information (optical flow) $v_t$ between the current frame $f_t$ and the previous reconstructed frame $\overline{f}_{t-1}$ is obtain through a lightweight DNN-based model proposed in \cite{ilg2017flownet}. To further improve the compression ratio, we encode, quantize and decode $v_{t}$ to obtain $\widehat{v}_{t}$. The specific network structure of the information codec (encoder and decoder) is provided in the appendix.
    
    \noindent\textbf{Step 4 Motion compensation.} 
    A motion compensation module is designed to obtain the predicted foreground $\widehat{f}_{t}$ based $\widehat{v}_{t}$ and previous reconstructed frame $\overline{f}_{t-1}$. This step will be described in detail in Sec.~\ref{sec:3.4}.
    
    \noindent\textbf{Step 5 Residual coding.} 
    The residual $r_{t}$ between the original foreground $f_{t}$ and the predicted foreground $\widehat{f}_{t}$ is obtained as $r_{t}$ = $f_{t}$ - $\widehat{f}_{t}$. Similarly, we use the same information codec as in Step 3 to encode, quantize, and decode $r_{t}$ to obtain $\widehat{r}_{t}$. Then we get the reconstructed foreground $\overline{f}_{t}$ = $\widehat{r}_{t}$ + $\widehat{f}_{t}$
    
    \noindent\textbf{Step 6 Two-stage decoding module.} 
    The composite module and Rec-Net make up a coarse-to-fine two-stage module. The composite module combines $\overline{f}_{t}$ with the background $\overline{b}_{t}$ to generate a composite frame $\widehat{x}_{t}$. The Rec-Net enhances the quality of composite frame $\widehat{x}_{t}$ by eliminating artifacts, blurring and boundaries and produces a visually more realistic frame $\overline{x}_{t}$. This step will be described in detail in Sec.~\ref{sec:3.5}.
    
    Step 1 is performed offline to extract foreground $f_t$ and background $b_t$ from $x_t$ for training, and then all networks in Step 2 - Step 6 are trained in an end-to-end manner. The detailed training strategy can be found in Section \ref{sec:4}.
    
    \subsection{Foreground Regions Proposal}\label{sec:3.2}
    First, the messy foreground points $p_t$ are extracted from the original frame $x_t$ through adaptive GMM. Second, morphological processing, such as thresholding, dilation, and open operation, is performed to remove noise. Third, if we compress background regions around foreground objects together when performing foreground compression, the transition between foreground and background in the final decoded frame will be more harmonious. Therefore, the simple bounding rectangle search \cite{cheng2008algorithm} is further applied to get regular rectangular regions. The above three steps are collectively expressed as the symbol $FP(\cdot)$. Finally, the foreground $f_t$ is obtained by the following formula:
    
    \begin{equation}
    m_t = FP(x_{t-1}) \cup FP(x_t)
    \end{equation}
    
    \begin{equation}
    f_t = m_t \odot x_t
    \end{equation}
    
    \noindent where $m_t$ is the foreground mask, $x_t$ is the original frame, and $f_t$ is the extracted foreground. $\cup$ is the AND-operation, and $\odot$ is the Hadamard product. 
    
    \begin{figure}[h]
        \begin{center}
            \includegraphics[width=1\linewidth]{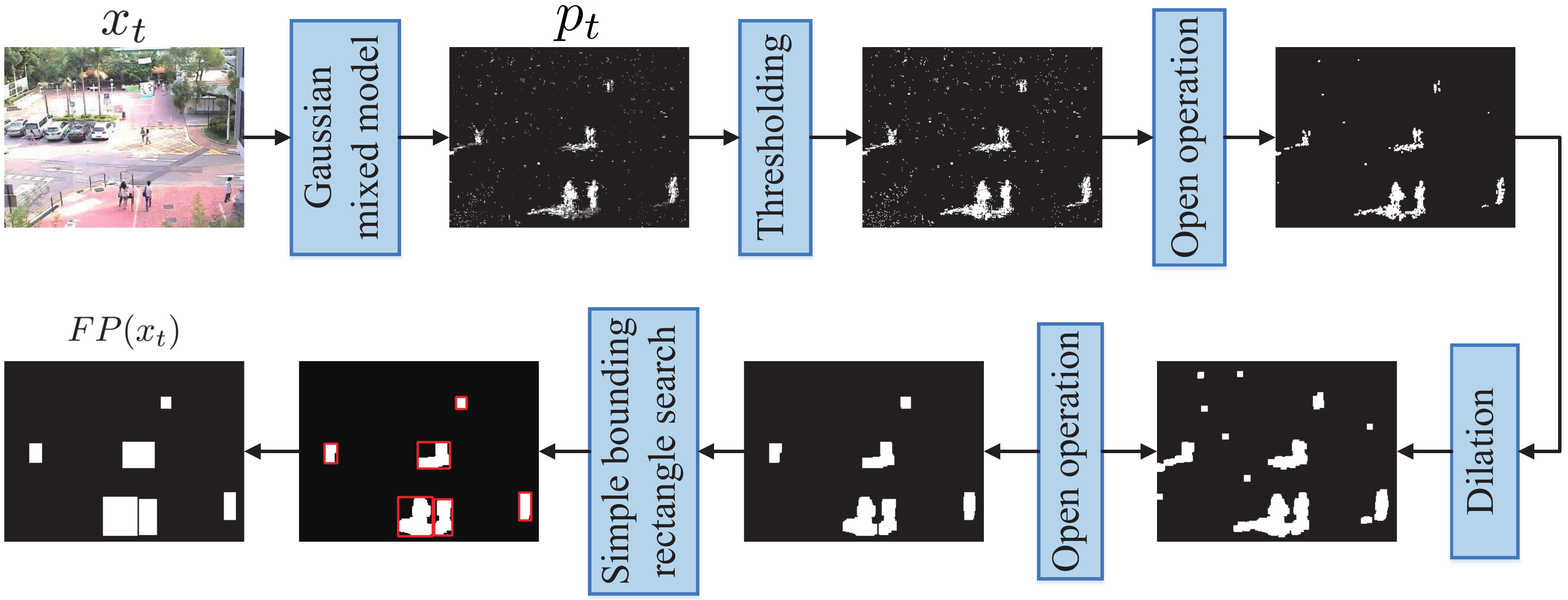}
        \end{center}
        \caption{Illustration of the processing $FP(\cdot)$ including adaptive Gaussian mixture model, morphological processing and simple bounding rectangle search.}
        \label{fig:2}
    \end{figure}
    
    \subsection{Adaptive Background Updating and Interpolation}\label{sec:3.3}
    Although the background of surveillance video is visually static, it actually has fluctuations in the brightness and color caused by the change of the lighting, weather, and time \cite{yang2018multi}. Therefore, we can select an initial background template, then update the template according to the changing intensity of background, and finally compute the remaining backgrounds by frame interpolation between adjacent templates. The background template can be updated at equal interval $l$, which means $b_{t}$, $b_{t+l}$, $b_{t+2l}$, \dots are selected as the background templates. However, it is not easy to determine a suitable fixed interval $l$. If $l$ is too large, the template may not respond to a fast and intense background change. If $l$ is too small, the compression ratio will be reduced, which is contrary to our original intention. To address it, an adaptive background template updating algorithm is proposed in this paper.
    
\begin{figure}[h]
        \begin{center}
            \includegraphics[width=1\linewidth]{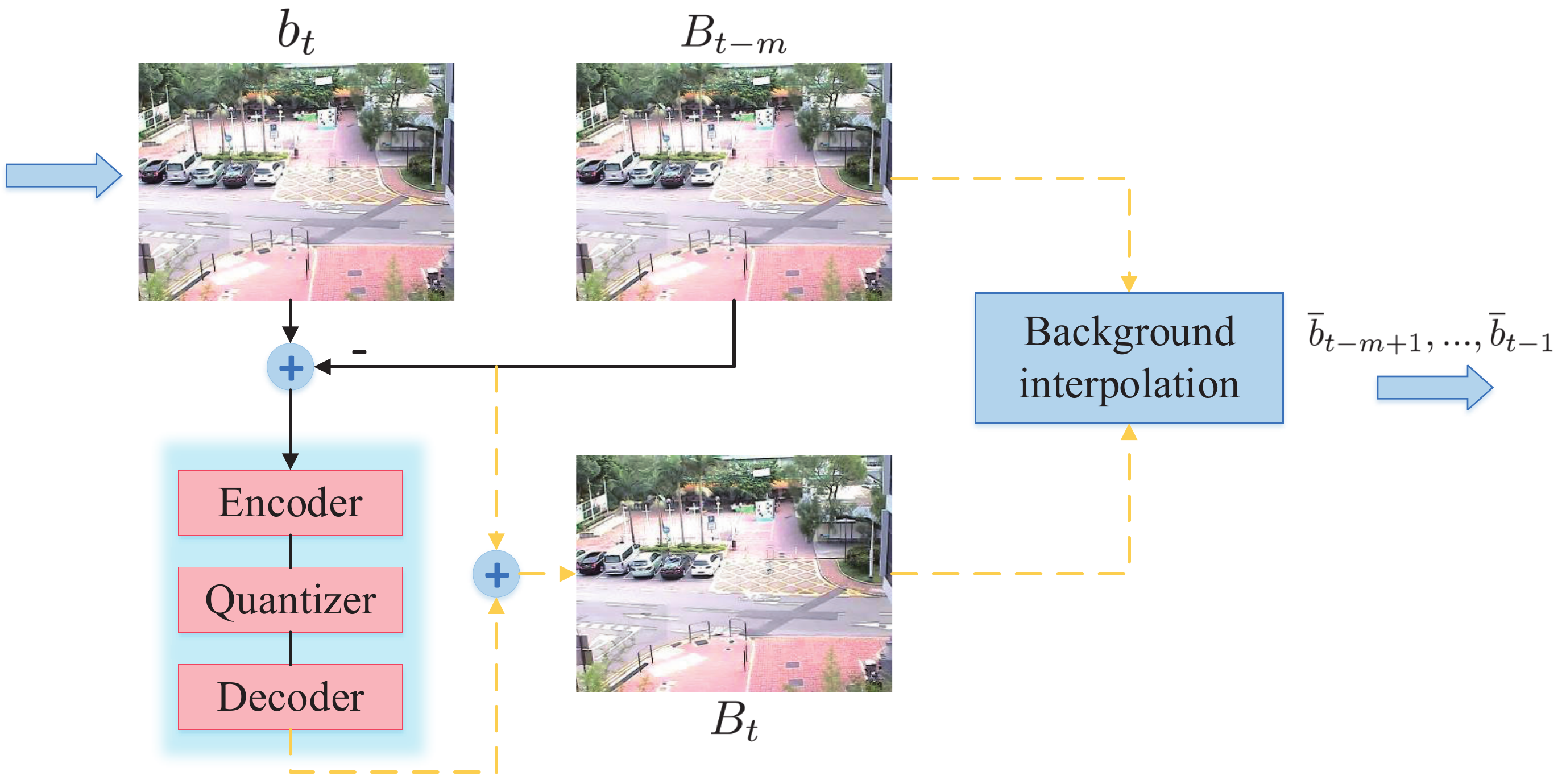}
        \end{center}
        \caption{Illustration of the adaptive background updating and interpolation.}
        \label{fig:3}
    \end{figure}

    First, the background at time $t-m$ is set as a background template $B_{t-m}$, and then the background candidate $b_{t}$ at time $t$ is compared with it. As shown in Fig.~\ref{fig:3}, if the MS-SSIM between $B_{t-m}$ and $b_t$ is smaller than a preset updating threshold $\gamma$ that controls the updating frequency of background, we perform residual coding on background $b_{t}$ as in Step 3 of Sec.~\ref{sec:3.1} to get new background template $B_{t}$. After that, the remaining backgrounds $\overline{b}_{t-m+1}, ..., \overline{b}_{t-1}$ between adjacent templates $B_{t-m}$ and $B_{t}$ can be obtained through linear interpolation, formulated as
    
    \begin{equation}
    \overline{\mathbf{b}}_{t-j}(i) = \mathbf{B}_{t-m}(i) + (\mathbf{B}_{t}(i) - \mathbf{B}_{t-m}(i)) \times \frac{m-j}{m}
    \end{equation}
    
    \noindent where $\overline{\mathbf{b}}_{t-j}$ denotes the reconstructed background at time $t-j$ and $\mathbf{B}_{t}$ denotes the background template at time $t$. Besides, $j$ $\in$ (\{0, 1, ..., m\}), $i$ enumerates all positions in $\overline{\mathbf{b}}$ and $\mathbf{B}$, and $m$ is the interval of two adjacent templates $\mathbf{B}_{t-m}$ and $\mathbf{B}_{t}$. In addition, we also use a more complex CNN-based frame interpolation method for background interpolation and present a trade-off between the two through experiments in Sec.~\ref{sec:5.5}. The results show that the CNN-based interpolation method improves the performance little but at the cost of a heavy calculation burden. Therefore, if not specifically mentioned, we use linear interpolation by default.
    
    \subsection{Motion Compensation}\label{sec:3.4}
\begin{figure}[h]
		\begin{center}
			\includegraphics[width=1\linewidth]{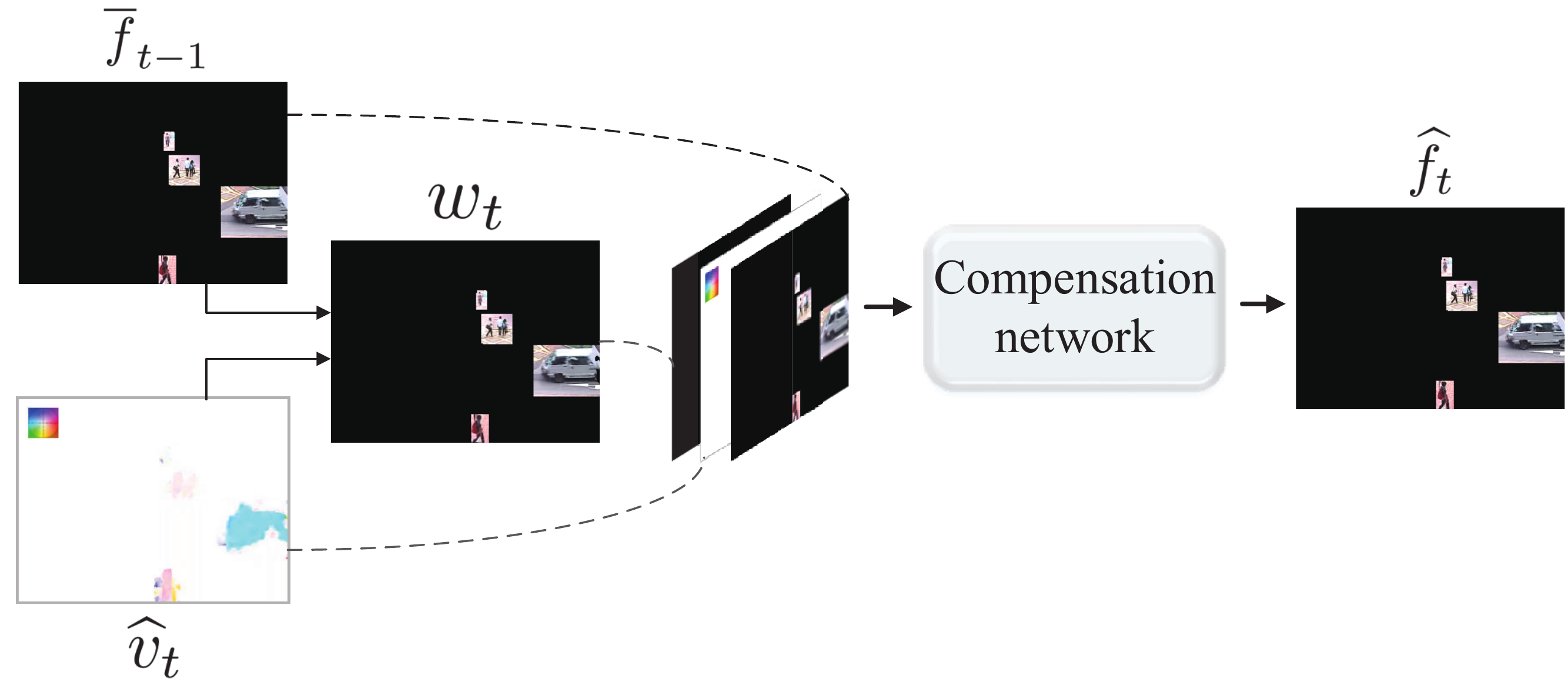}
		\end{center}
		\caption{Illustration of the motion compensation module.}
		\label{fig:4}
	\end{figure}

    As shown in Fig.~\ref{fig:4}, the effect of motion compensation is essentially to obtain the predicted foreground $\widehat{f}_{t}$ based on the previous reconstruction foreground $\overline{f}_{t-1}$ and motion information $\widehat{v}_{t}$. Our method uses optical flow as motion information to achieve pixel-level motion compensation, which has higher accuracy and flexibility than conventional block-level motion compensation.
    
    First, the wrapped foreground $w_{t}$ is obtained with the following remapping formula:
    
    \begin{equation}
    \overline{f}_{t-1}(x, y) = w_{t}(x + v_{x}, y + v_{y})
    \end{equation}

    \noindent where $(x, y)$ is a point in $\overline{f}_{t-1}$ and $(v_{x}, y_{y})$ represents the motion information of point $(x, y)$ from time $t-1$ to time $t$. Then we contact the $\overline{f}_{t-1}$, $w_{t}$ and $\widehat{v}_{t}$ along the channel axis into a feature map and input it into the compensation network (Com-Net) to obtain the predicted foreground $\widehat{f}_{t}$. The specific structure of the Com-Net is provided in the appendix.
    
	\begin{figure*}[h]
		\begin{center}
			\includegraphics[width=0.9\linewidth]{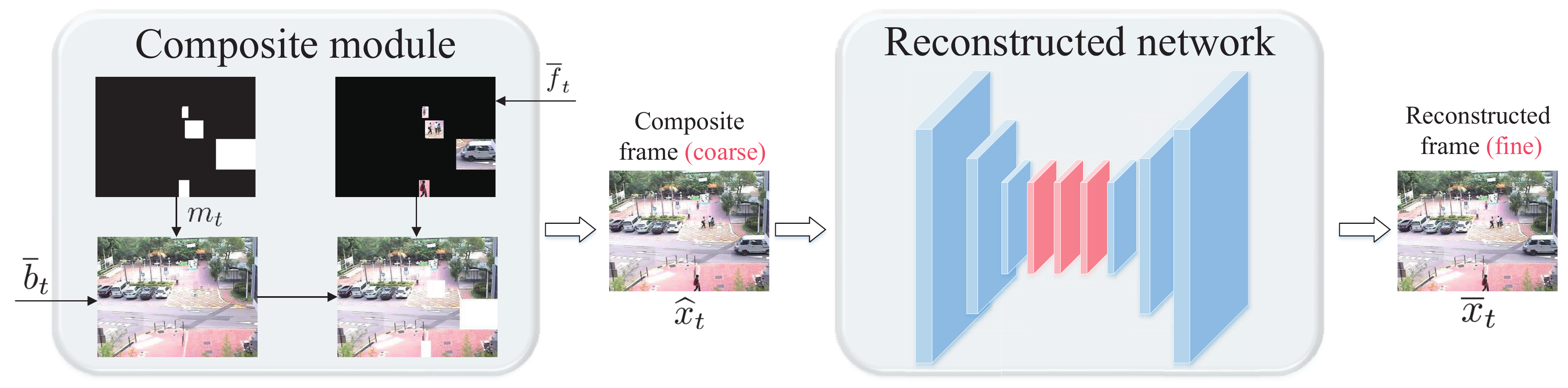}
		\end{center}
		\caption{Illustration of a coarse-to-fine, two-stage module, which is composed of a composite module and a reconstruction network (Rec-Net). The composite module embeds the reconstructed foreground $\overline{f}_{t}$ in the corresponding position of the reconstructed background $\overline{b}_{t}$. The quality of the composite frame $\widehat{x}_{t}$ is enhanced by the Rec-Net to eliminate distortion such as artifacts, boundaries and blurs. The blue planes in Rec-Net represent the convolution and transposed convolutional layers, and the red planes represent the residual blocks.}
		\label{fig:5}
	\end{figure*}

    \subsection{Coarse-to-fine Two-stage Decoding Module}\label{sec:3.5}
    As shown in Fig.~\ref{fig:5}, the composite module is used to combine reconstructed foreground $\overline{f}_{t}$ with reconstructed background $\overline{b}_{t}$ to produce a composite frame $\widehat{x}_{t}$. Specifically, it is to replace the corresponding position of reconstructed background $\overline{b}_{t}$ with reconstructed foreground $\overline{f}_{t}$ according to the foreground mask $m_{t}$. However, the foreground and background in the composite frame $\widehat{x}_{t}$ may not be well merged, and there may be transition boundaries and inconsistencies. 
    
    The quality of the composite frame $\widehat{x}_{t}$ is enhanced through the Rec-Net to make foreground and background composited harmoniously. As shown in Fig.~\ref{fig:5}, the Rec-Net is composed of a stack of three convolutional layers, three residual blocks, and three transposed convolutional layers. The specific network structure of the Rec-Net is provided in the appendix.

    \section{Implementation}\label{sec:4}
    \subsection{Quantization}
    The quantization approach proposed in \cite{theis2017lossy} is adopted in our work to quantize motion $v_t$ and residual $r_t$ into $v^{q}_t$ and $r^{q}_t$. Specifically, given centers $C_L$ = \{0, 1, 2, ..., $2^L$-1\}, the nearest neighbor assignments is applied to compute:
    
    \begin{equation}
    \widehat{q}=Q(\omega)=\arg\min\limits_{j}|\omega-c_j|
    \end{equation}
    
    \noindent but rely on (differentiable) soft quantization 
    
    \begin{equation}
    \overline{q}=\sum_{j=1}^{L} \frac{\exp \left(-\sigma\left\|w-c_{j}\right\|\right)}{\sum_{l=1}^{L} \exp \left(-\sigma\left\|w-c_{l}\right\|\right)} c_{j}
    \end{equation}
    
    \noindent to compute gradients during the backward pass.
    
    \subsection{Training Strategy}
    The parameters of our model can be trained in an end-to-end manner. First, we initialize the GMM with a learning rate of 0.005, an initial variance of 15, and a variance threshold of 16. GMM is trained based on the Expectation-Maximization (EM) algorithm with 200 frames. In our implementation, the background template obtained after GMM initialization can be used to guide the compression of these initial frames. Furthermore, considering that our compression system is specifically designed for surveillance scenarios, the impact of the initialization delay (less than 10 $s$) on performance is almost negligible in practical applications since the surveillance videos are usually working for 24 hours. Then Step 1 is performed with the pretrained GMM to obtain foreground $f_t$ and background $b_t$ in an offline manner. 
    
    Finally, we perform the end-to-end training on all networks including in Step 2-Step 6. It worth noting that the template updating and background interpolation in Step2 is only used during testing because background updating and interpolation are not trainable. During training, every frame of the background is processed by the background residual compression network. The purpose of the training is to minimize the following loss:
    
    \begin{equation}
    L=\alpha E\left[d\left(x_{t}, \overline{x}_{t}\right)\right] + \beta E\left[d\left(f_{t}, \overline{f}_{t}\right)\right] + \theta H(v^{q}_t, r^{q}_t)
    \end{equation}
    
    \noindent The first term denotes the distortion between $x_t$ and $\overline{x}_{t}$ and we use Mean Square Error (MSE) in our implementation, that is, $d=(x_{t}-\overline{x}_{t})^2$. To focus on the reconstruction of foreground regions, we add a second term $E\left[d\left(f_{t}, \overline{f}_{t}\right)\right]$ that denotes the distortion between $f_t$ and $\overline{f}_{t}$. The effects of the above two items are controlled by the parameters $\alpha$ and $\beta$, respectively. Instead of encoding all positions with the same number of bits, we added a third term $H(v^{q}_t, r^{q}_t)$ to optimize both the bit rate and distortion, where $H(\cdot)$ represents the number of bits used for encoding $v^{q}_t$ and $v^{q}_t$. Following \cite{wu2018video,mentzer2018conditional}, a 3D-CNN is used as the context model to obtain the probability of each bit and then $v^{q}_t$ and $r^{q}_t$ are encoded with bit probability and adaptive arithmetic coding. This idea was first elaborated in \cite{mentzer2018conditional} for image compression. Since then it has also been used as an entropy term in \cite{wu2018video} for video compression and has achieved good results.The reasons why the bits for the background are not included are as follow: (1) The quality of the background template compression must be ensured, because a low-quality background template will affect the performance of all backgrounds obtained by interpolation. Therefore, the background compression should give priority to the quality instead of the rate to minimize the distortion; (2) Compared with foreground bits, background bits account for only a small fraction of the total bits. Thus, the total number of bits does not increase much even without rate optimization for background bits.
    
    \subsection{Bitstream Storage}
\begin{figure}[htb]
        \begin{center}
            \includegraphics[width=0.8\linewidth]{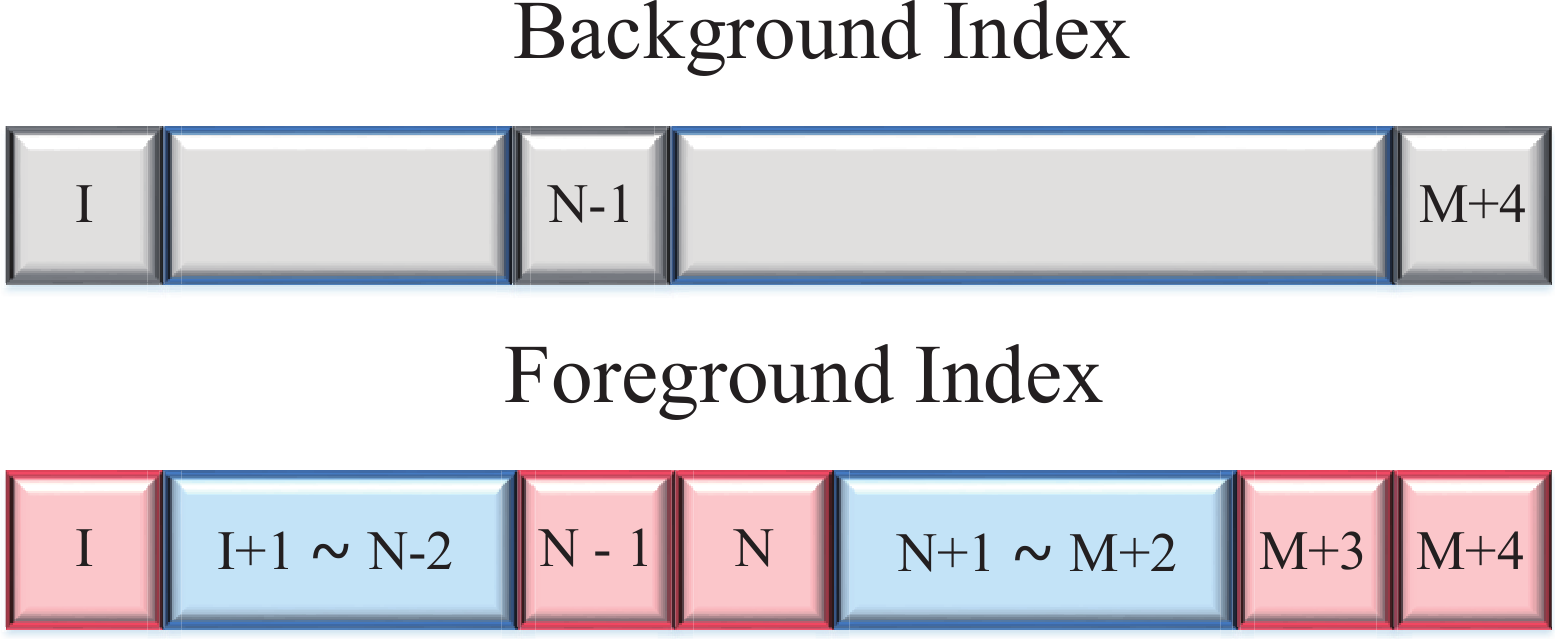}
        \end{center}
        \caption{Illustration of the background and foreground index. The letters ``$I, N, M$" represent the frame number. For example, ``$N-1$" represents the $(N-1)th$ frame, and ``$N + 1\sim M + 1$" represents a frame segment range from $N+1$ to $M+1$.}
        \label{fig:6}
    \end{figure}

    In our method, the compression bitstreams of the foreground and background are indexed to facilitate retrieval. Specifically, as shown in Fig.~\ref{fig:6}, the compression bitstreams of the background templates are stored along with the background index of their frame number. Also, the foreground index indicates the corresponding frame number of each foreground and the frame range of non-foreground segments.
    
    \section{Experiments}\label{sec:5}
    \subsection{Experiment Setup}
    \subsubsection{Dataset}
    The following two publicly available video datasets are used in our experiments: CUHK Square Dataset \cite{wang2012transferring} (contains a one-hour surveillance video) and EWAP Dataset \cite{pellegrini2009you} (contains two ten-minute surveillance videos). The above three videos contain many different sizes of objects, such as cars, people and bicycles. What’s more, there are changes in brightness with light and weather as in a natural environment. To remove artifacts induced by the previous compression, each of the video frames is cropped and resized to 320 $\times$ 240 $\times$ 3 before training. Each of these three videos is divided into 1-minute clips to form a new dataset. For each video, 70\%, 20\%, and 10\% of the clips were used for training, validation, and testing, respectively. The training clips from the three videos are mixed together to form a new training set, which is used to train our proposed algorithm and the baseline algorithm. The obtained models are then fine-tuned and evaluated on the validation and test sets of the three videos separately. Besides, the standard HEVC Dataset \cite{sullivan2012overview} is also used as a test set to compare with the performance of different methods. However, since our method is specifically designed for surveillance scenarios, only six videos with the static background (Traffic, Cactus, BasketballDrill, Vidyo1, Vidyo3, Vidyo4) are selected for testing.
    
    \subsubsection{Evaluation Standard}
    The key to compression is to find a balance between distortion and compression ratio, which are important basis for us to measure the performance of a compression algorithm. In this paper, PSNR and MS-SSIM are used to measure compression distortion. Besides, the compression ratio is measured by the average bits required to encode each pixel per frame ($bpp$).
    
    \subsubsection{Parameter Settings}
    The weights $\alpha$, $\beta$ and $\theta$ of three loss components mentioned in Eq.~7 are set to 1, 16 and 0.1, respectively. In addition, the threshold parameter $\gamma$ related to background updating frequency is set to 0.98. In the paper, foreground motion, foreground residuals, and background residuals share the same quantization parameter $L$, which is set to different values to obtain various compression ratios and performance. If not specified, $L$ is set to the minimum value 1 by default, which performs binary quantization on $v_t$ and $r_t$, which has been proved to be effective in compression methods such as \cite{agustsson2018generative,li2018learning}. During the training process, the model is iteratively trained 128 times on the dataset. The initial learning rate is set to $2 \times 10^{-4}$, and after 64 iterations, the learning rate is changed to $2 \times 10^{-5}$.

    \subsection{Adaptive Background Template Updating Analysis}
    The background in the surveillance video has fluctuations, and there are minor changes in the brightness and color of the background between two adjacent frames. The handcraft background updating frequency often fails to respond to changes in the natural environment in a timely manner, so an adaptive template updating algorithm is adopted in this paper. A total of 900 frames are selected to calculate MS-SSIM between background extracted from each frame and its previous background template according to the background template updating algorithm as described in Sec.~\ref{sec:3.3}. When MS-SSIM is smaller than the background updating threshold $\gamma$, which indicates that the background has changed a lot, possibly because the sun is blocked by the cloud or at dusk.
    
\begin{figure}[h]
    	\begin{center}
    		\includegraphics[width=1\linewidth]{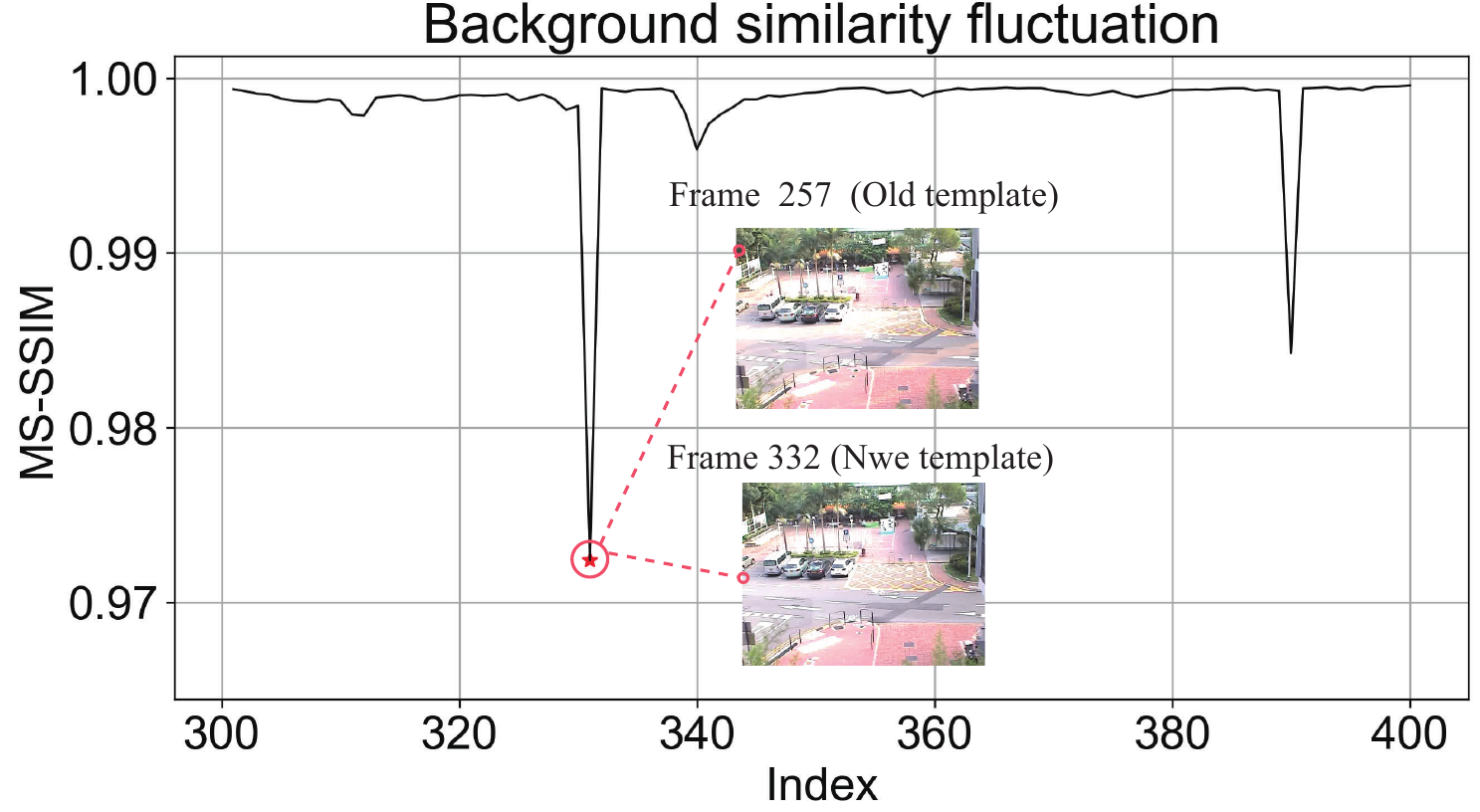}
    	\end{center}
    	\caption{Illustration of background fluctuations (Frame 300-400). The moment circled is when the background changes a lot and needs to be updated.}
    	\label{fig:7}
    \end{figure}

    As shown in Fig.~\ref{fig:7}, when the weather turns cloudy, MS-SSIM is smaller than 0.98, which is the moment when the background template needs to be updated. The update frequency of the background template is controlled by the threshold of $\gamma$, which affects the compression ratio of background compression.
    
    \subsection{Comparison of Results}
    Our method is compared with conventional algorithms H.264, H.265 as well as BCBR \cite{zhang2013background}, a method specially designed for surveillance video compression. We follow the settings in \cite{lu2019dvc} to get frames compressed by H.264 and H.265, and the low delay coding structure is applied in the paper. To avoid tedious notation, when we mention H.264 and H.265, we are referring to use x264 and x265 as the encoder. Instead, the implementation with HM 16.6 as the encoder is denoted as ``HEVC/HM". The official implementation and HM \textit{encoder\_lowdelay\_main.cfg} default profile (except QP is set to 32) are used in our experiments, and the filename, resolution, frame rate and other parameters are adjusted according to the specific video. In addition, the first and the latest end-to-end deep compression method KFI \cite{wu2018video} and DVC \cite{lu2019dvc} are also included in the comparison. Besides, we compare our method with post-processing algorithm MFQE \cite{yang2018multi} which is used to optimize the performance of H.265. The above three DNN-based schemes are implemented based on the Open Source Code and test model provided on Github \cite{RN1,RN12,RN123}, and then used as the baseline for comparison. 

\begin{figure}[h]
   		\begin{center}
   			\includegraphics[width=1\linewidth]{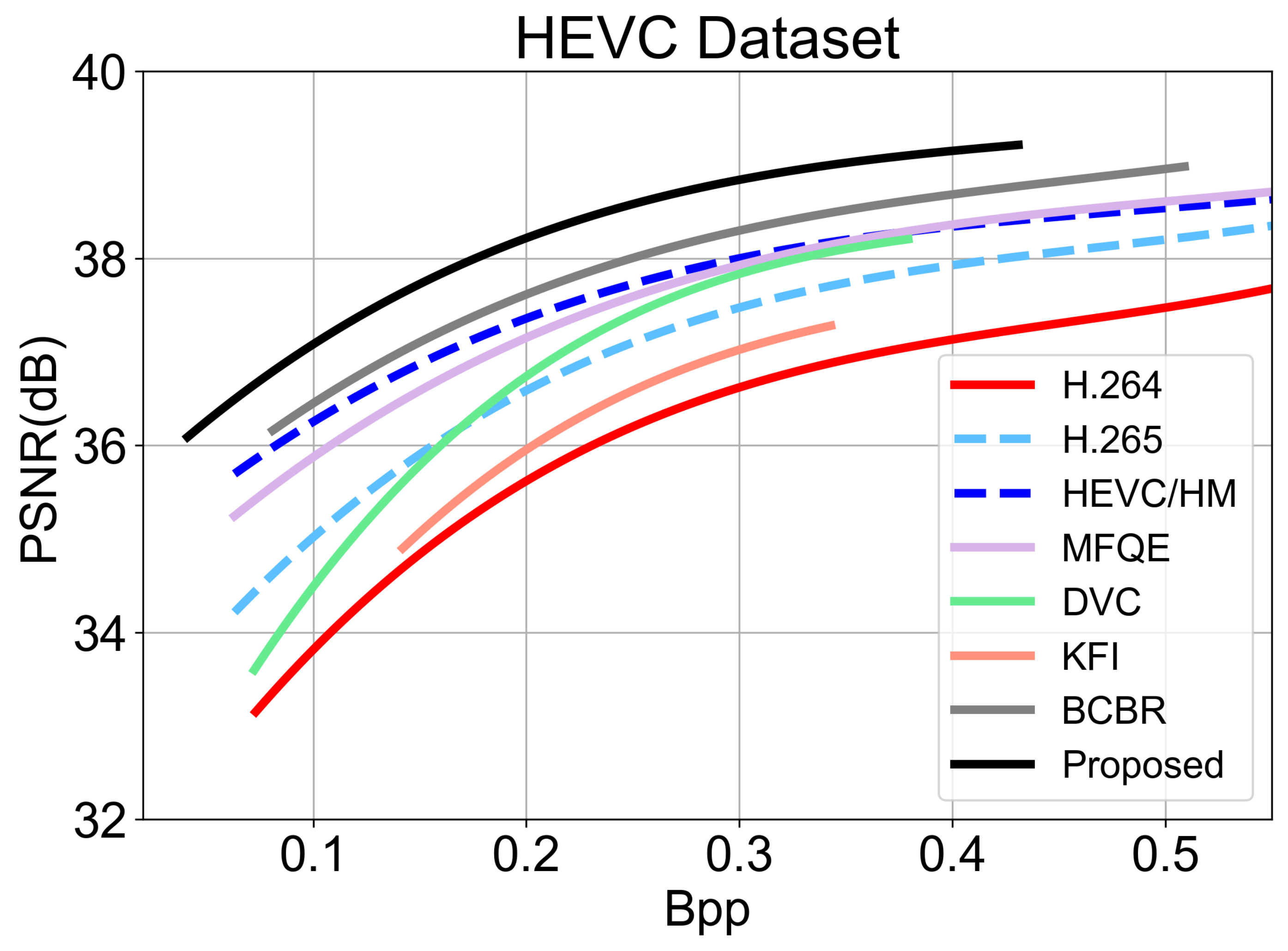}
   		\end{center}
   		\caption{Illustration of the comparison of our method with other conventional and DNN-based video compression methods.}
   		\label{fig:8}
   	\end{figure}

    As shown in Fig.~\ref{fig:8}, when evaluated by PSNR, our method outperforms other methods. The performance of KFI is close and slightly better than H.264, but much worse than H.265. The performance of DVC is slightly worse and slightly better than H.265 at low $bpp$ and high $bpp$, respectively. Although BCBR is somewhat better than HEVC/HM overall and MFQE improves the performance of H.265 by 0.5-1 dB, but there is still a big gap between their performance and ours. For example, to achieve the same PSNR (36 dB), our method requires 73.5\% less $bpp$ than H.265. Even when compared with HEVC/HM, which is more advantageous in surveillance scenarios, our method still performs better, i.e., 49.75\% less $bpp$ to achieve the same PSNR (36 dB). Furthermore, we found the two DNN-based end-to-end methods are not as good in performance as the conventional method BCBR, because they're not specifically designed for surveillance scenarios. Besides, we can see that our method has a greater advantage than other methods at low $bpp$. %As $bpp$ rises, the background compression algorithm gradually becomes an important factor for limiting performance improvement. Fortunately, the background compression algorithm is replaceable in our work. With the advent of algorithms with better performance, the performance of our method at high $bpp$ can be further improved. 
   
   	\begin{figure*}[t]
   		\begin{center}
   			\includegraphics[width=0.3\linewidth]{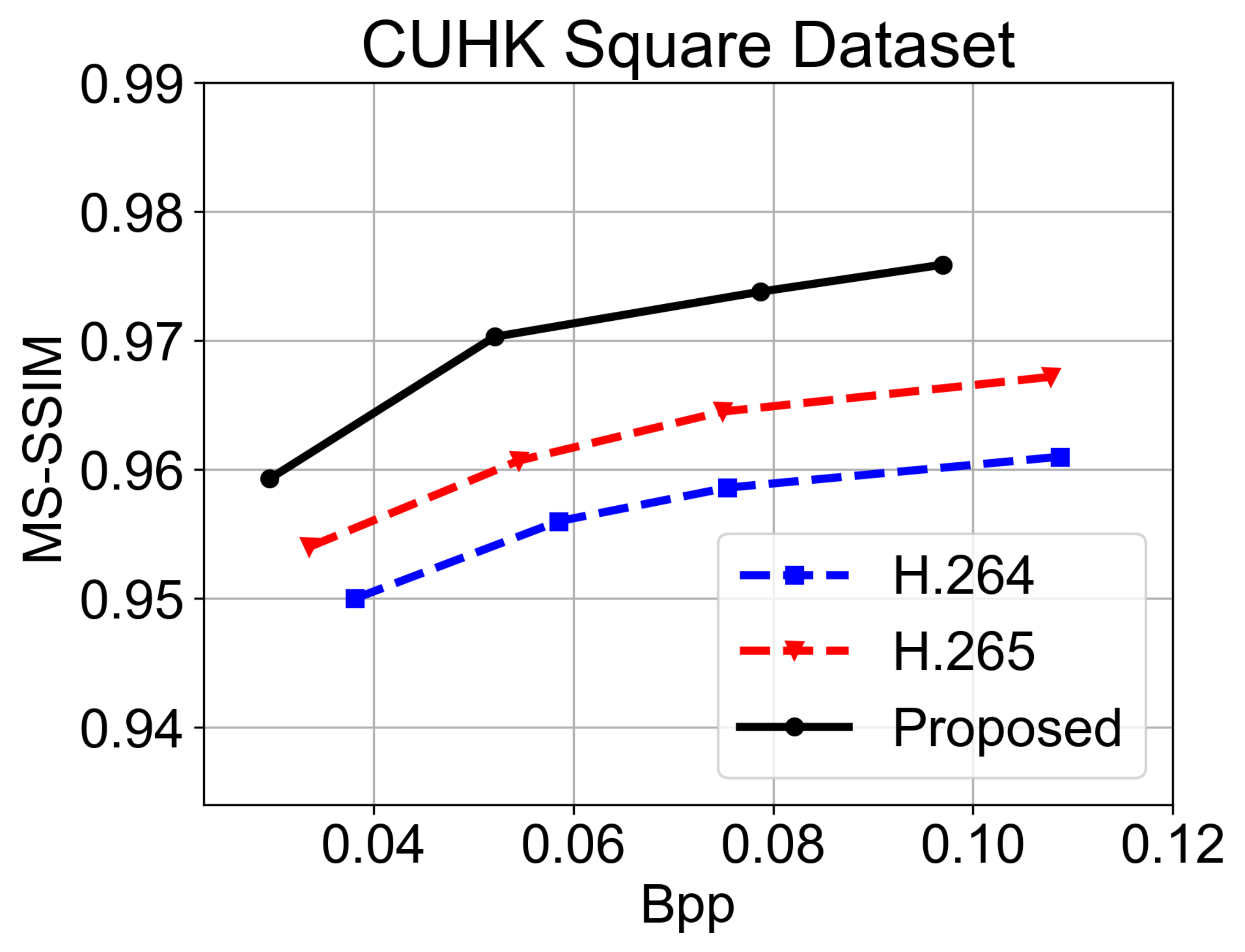}
   			\includegraphics[width=0.3\linewidth]{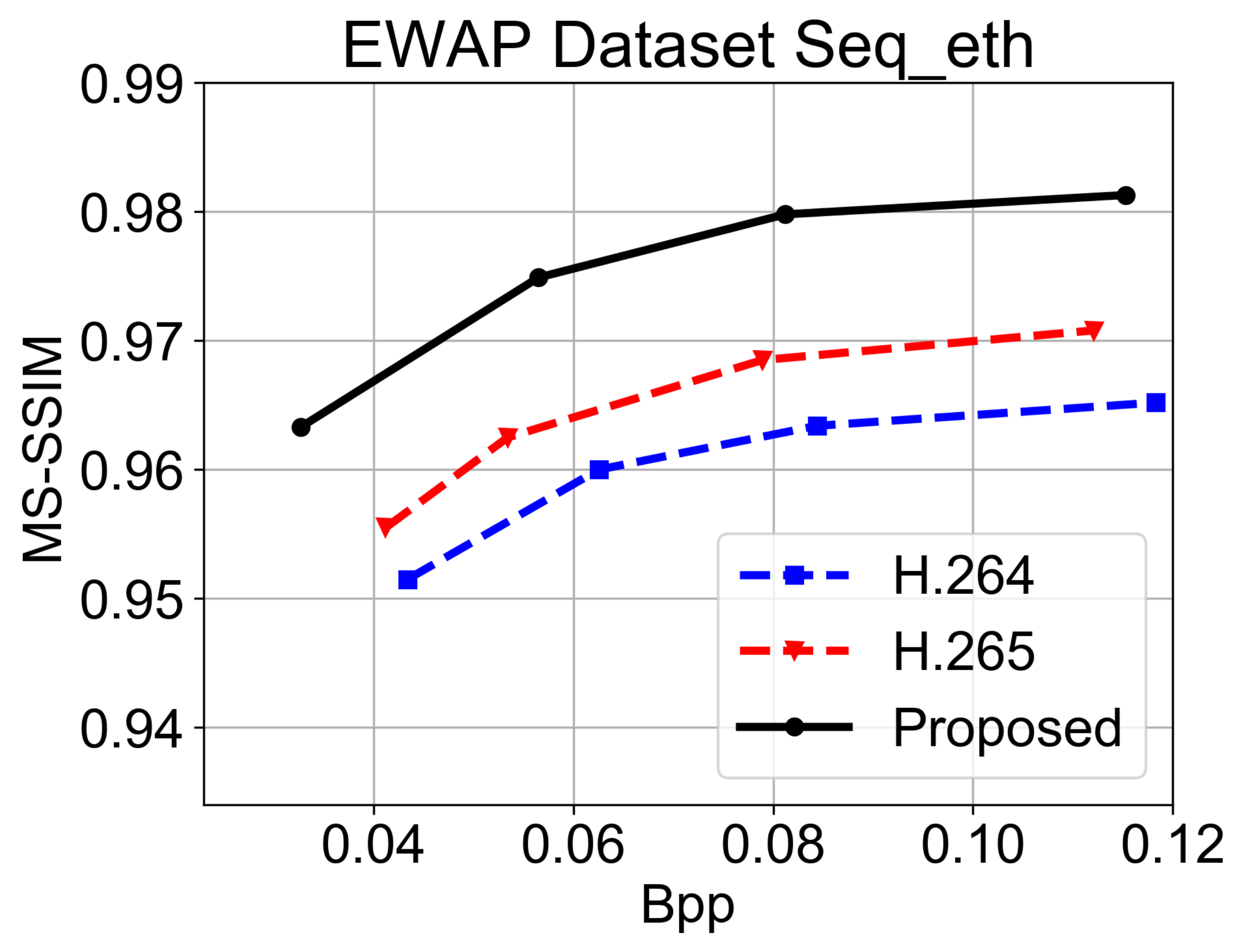}
   			\includegraphics[width=0.3\linewidth]{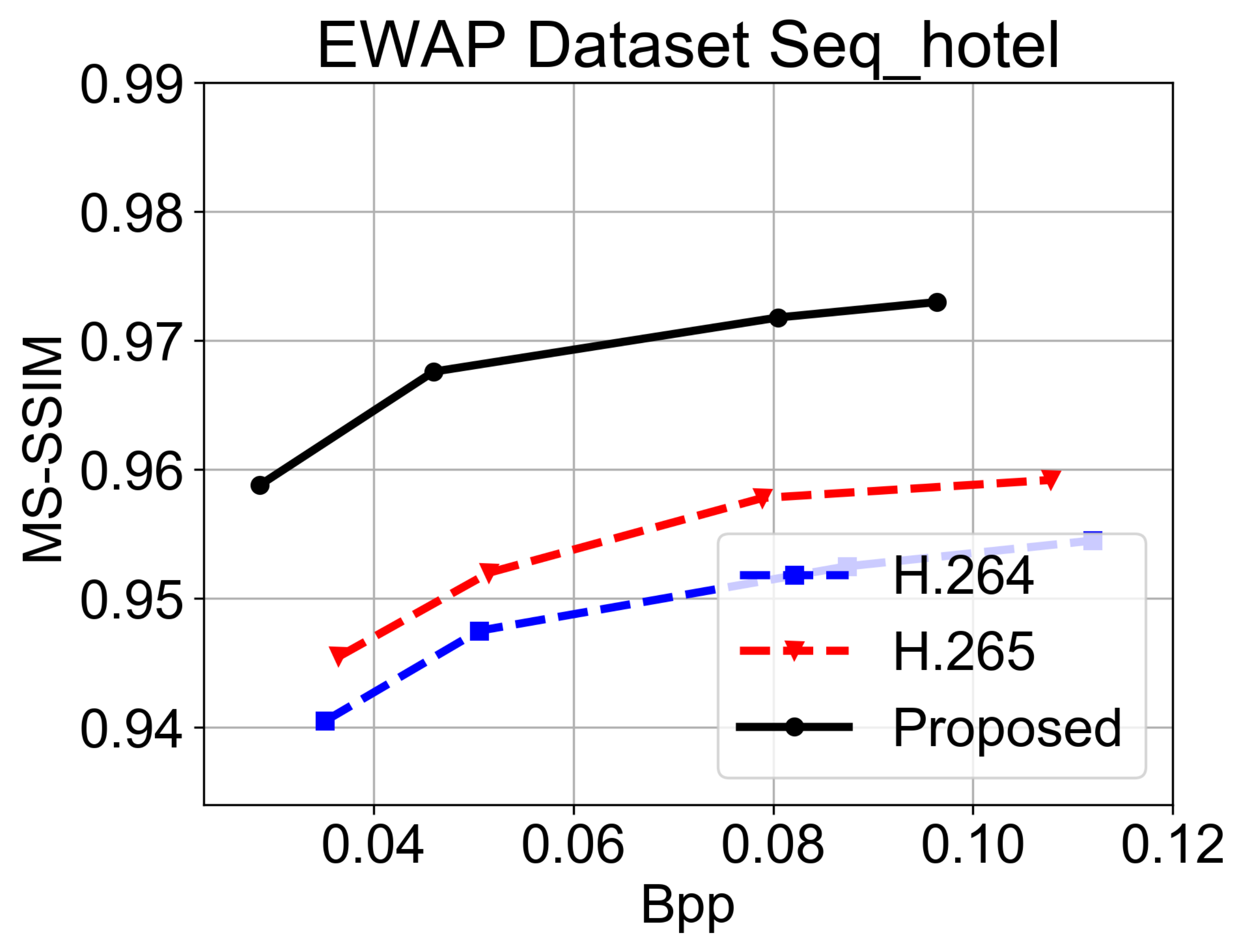}
   			\includegraphics[width=0.3\linewidth]{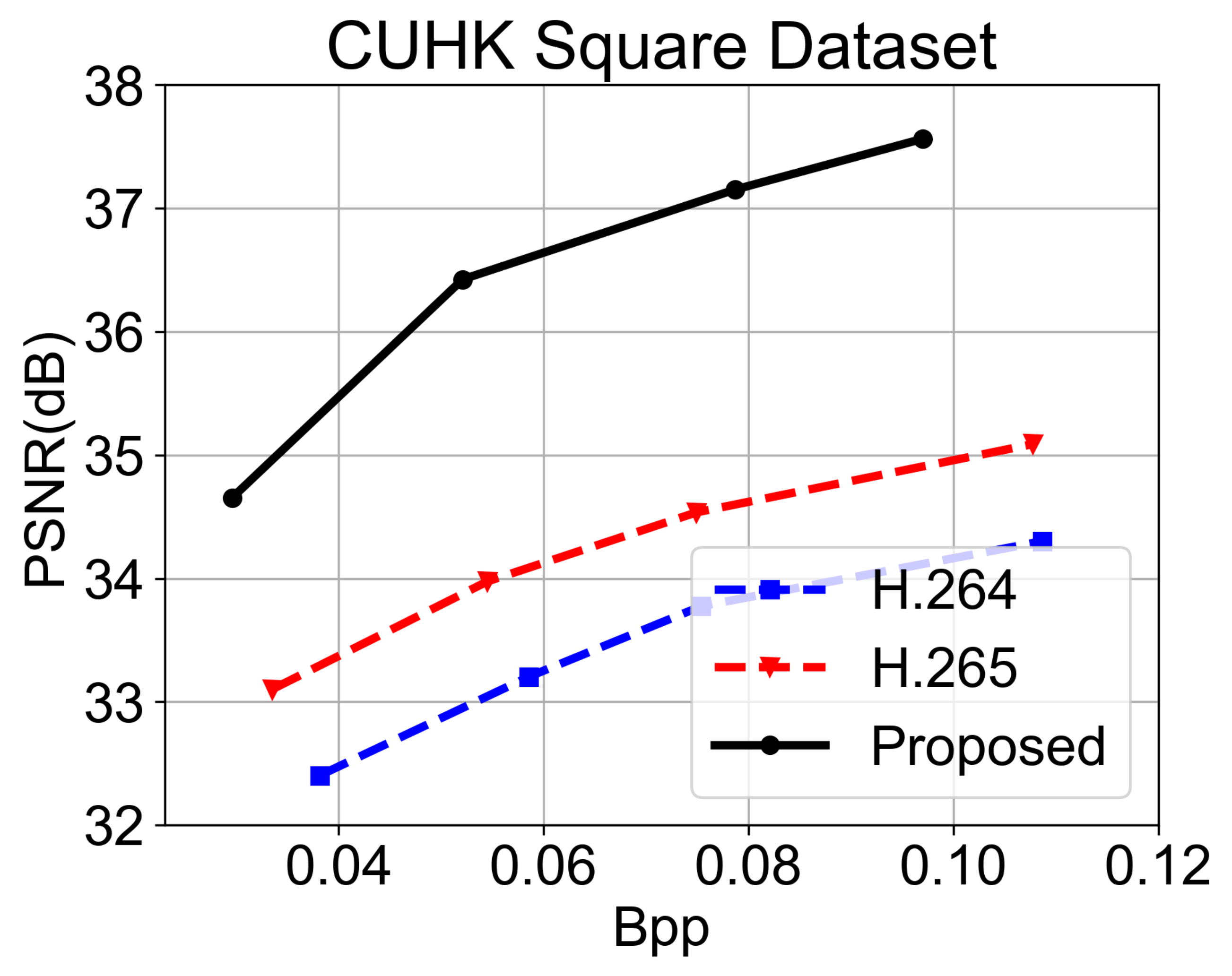}
   			\includegraphics[width=0.3\linewidth]{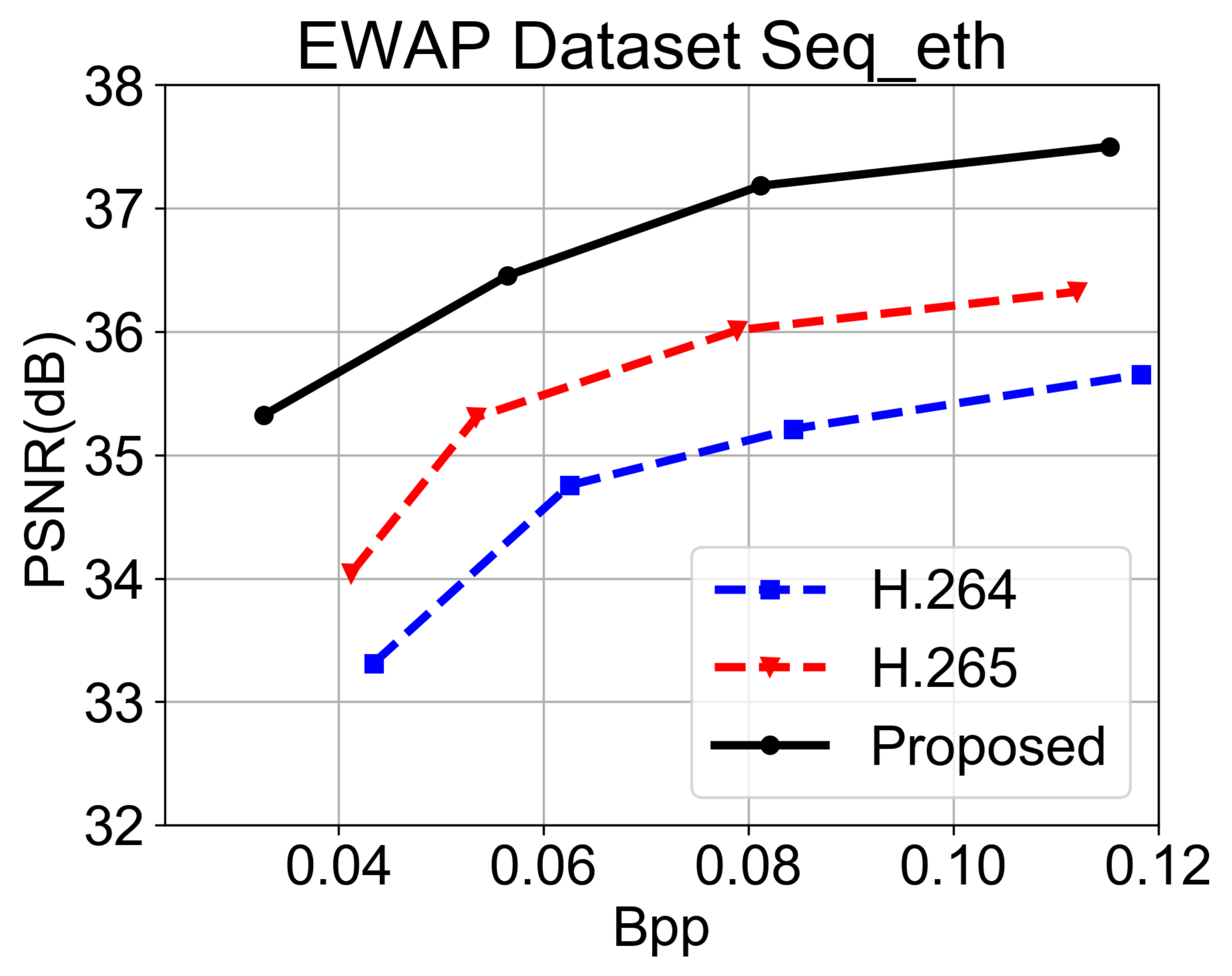}
   			\includegraphics[width=0.3\linewidth]{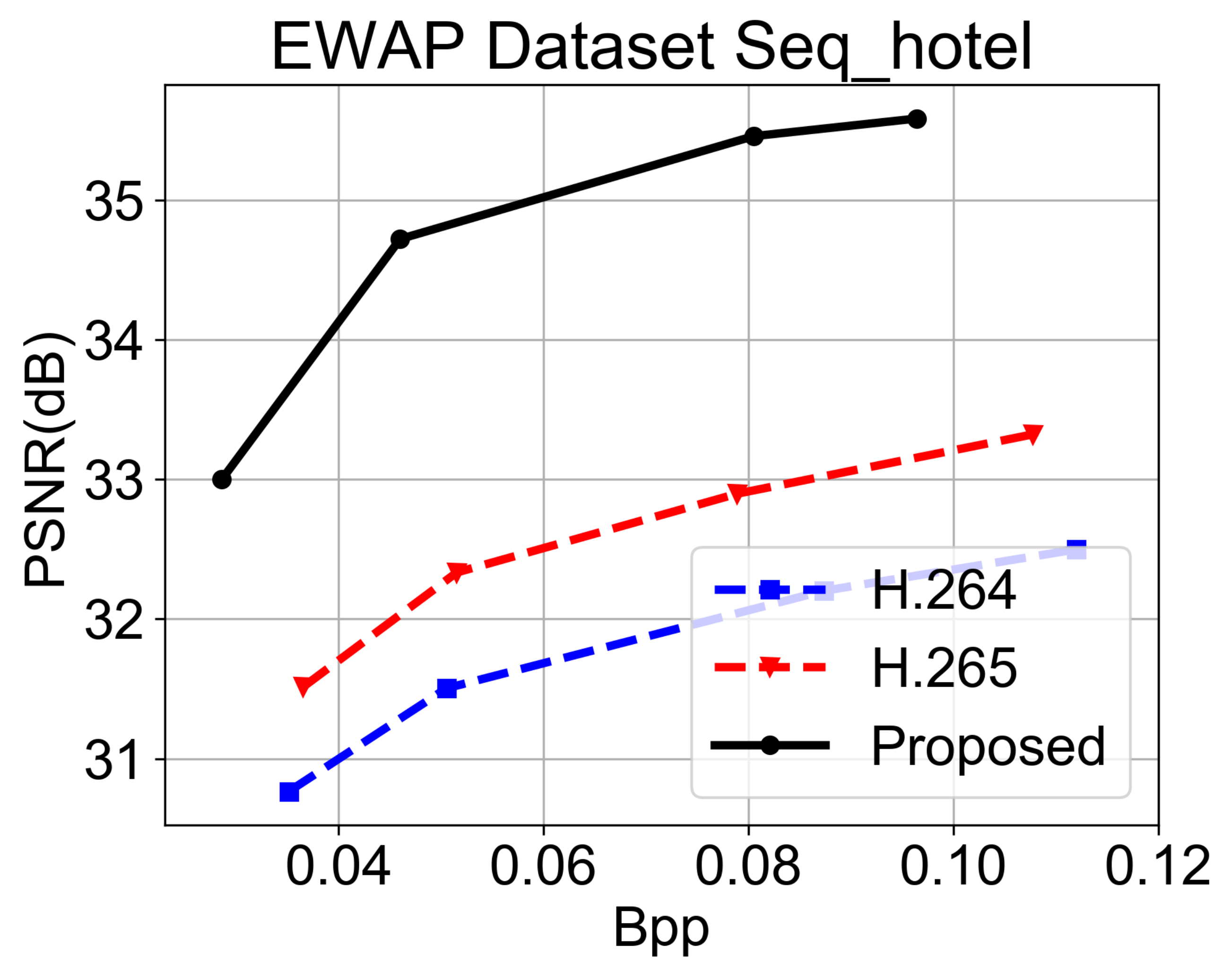}
   		\end{center}
   		\caption{Illustration of the comparison of results on the CUHK Square Dataset and EWAP Dataset.}
   		\label{fig:9}
   	\end{figure*}
   	
    In addition to the above comparisons, the model is also tested on the CUHK Square Dataset and EWAP Dataset and compared with the compression results of H.264 and H.265. The comparison results are shown in Fig.~\ref{fig:9}. The compression performance of our method is much better than H.264 and H.265 in both MS-SSIM and PSNR. For example, our proposed method needs 71.3\% and 46.5\% less $bpp$ to achieve the same MS-SSIM (0.96) than conventional algorithms H.264 and H.265 on the CUHK Square Dataset, respectively.
    
    In our framework, the number of bits assigned to Background Residuals ($BR$), Foreground Residuals ($FR$) and Foreground Motion Estimation ($FMV$) depends on the total number of bits used for compression. Therefore, our $FMV$ and $FR$ refer only to the foreground region, while the $MV$ reported in DVC refers to the full frame. To reflect this point, we present the bits percentages of background residuals, foreground residuals, and foreground motion information at different bits on the CUHK Square Datasets in Table.~\ref{table:3}. As shown in table, the ratio of bits assigned to background residuals is relatively small and stable across different $bpp$. As $bpp$ increases, the ratio of bits for foreground residuals gradually decreases, while the ratio of bits for foreground motion information increases. This is mainly because more bits are available for motion estimation at high $bpp$, enabling better preservation of motion information. Therefore, the prediction frames obtained by motion compensation are also more accurate, resulting in a higher ratio of zero in the obtained foreground residuals, and thus reducing the corresponding ratio of bits. With a context-based rate optimization strategy, the percentage of bits used for foreground residual compression will also be reduced.

    \linespread{1.2}
    \begin{table}[htbp]
        \centering
        \caption{Bit Allocation Ratio}
        \label{table:3}
        \begin{tabular}{l|c|c|c|c}
            \hline
            & 0.085 ($bpp$) & 0.174 ($bpp$) & 0.264 ($bpp$) & 0.352 ($bpp$) \\
            \hline
            BR & 6.53\% & 5.88\% & 6.02\% & 5.98\% \\
            \hline
            FR & 53.65\% & 51.76\% & 49.32\% & 44.54\% \\
            \hline
            FMV & 39.82\% & 42.36\% & 44.66\% & 49.48\% \\
            \hline        
        \end{tabular}
    \end{table}
    
    \subsection{Ablation Experiment}\label{sec:5.4}
\begin{figure}[h]
        \begin{center}
            \includegraphics[width=1\linewidth]{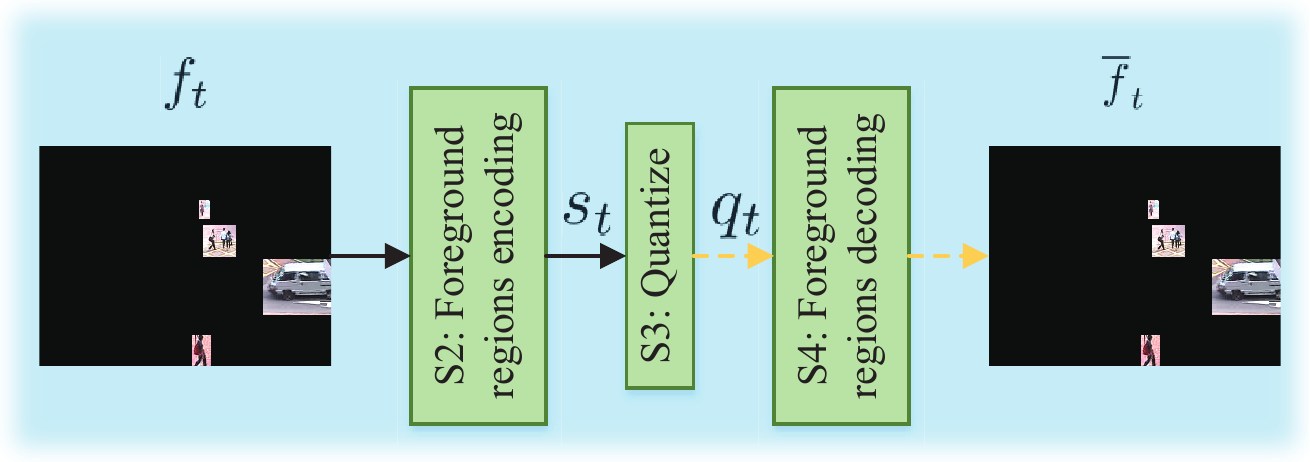}
        \end{center}
        \caption{Illustration of an autoencoder scheme that encodes and decodes the foreground directly.}
        \label{fig:10}
    \end{figure}

\begin{figure*}[h]
		\begin{center}
			\includegraphics[width=0.46\linewidth]{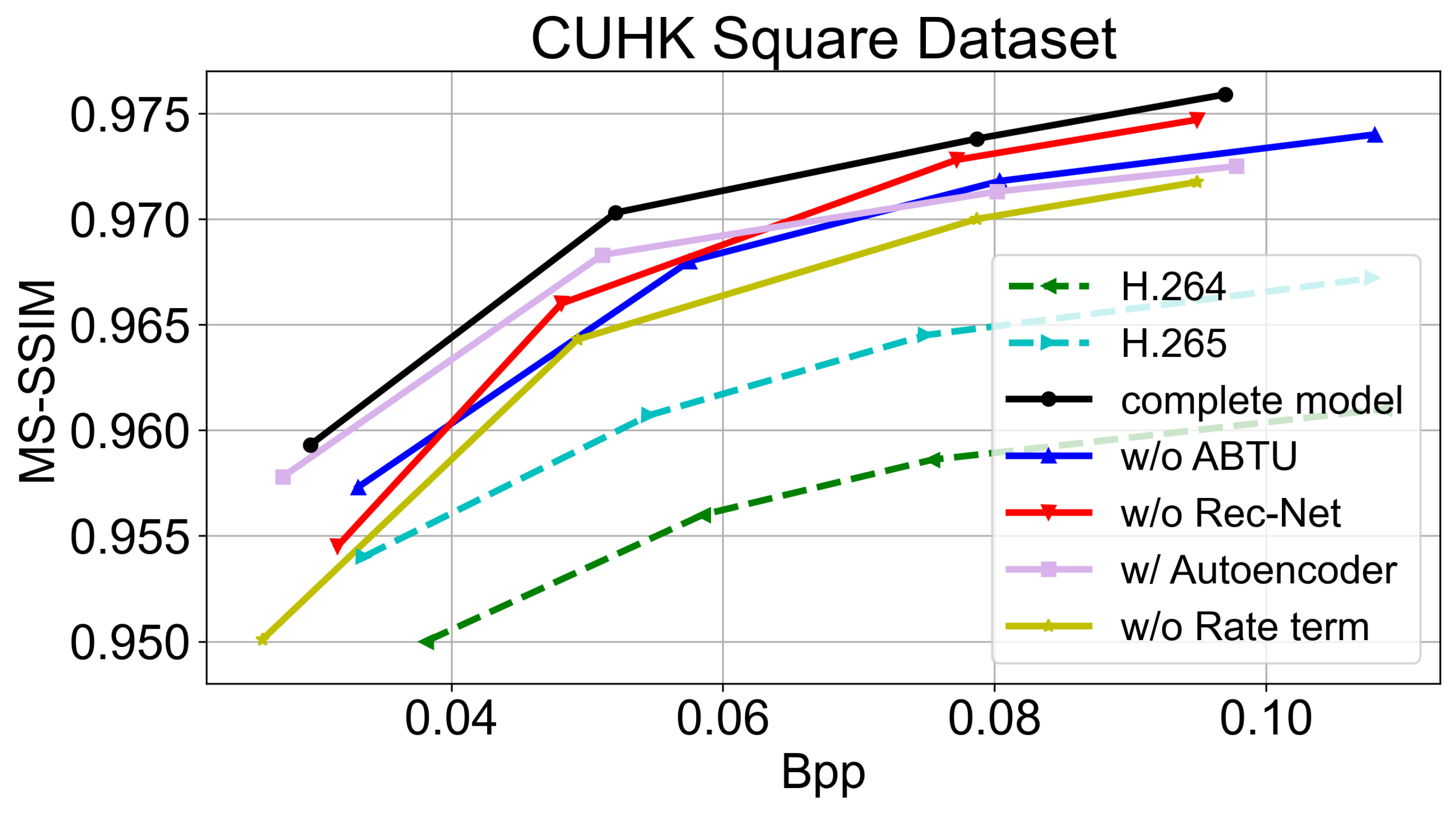}
			\includegraphics[width=0.46\linewidth]{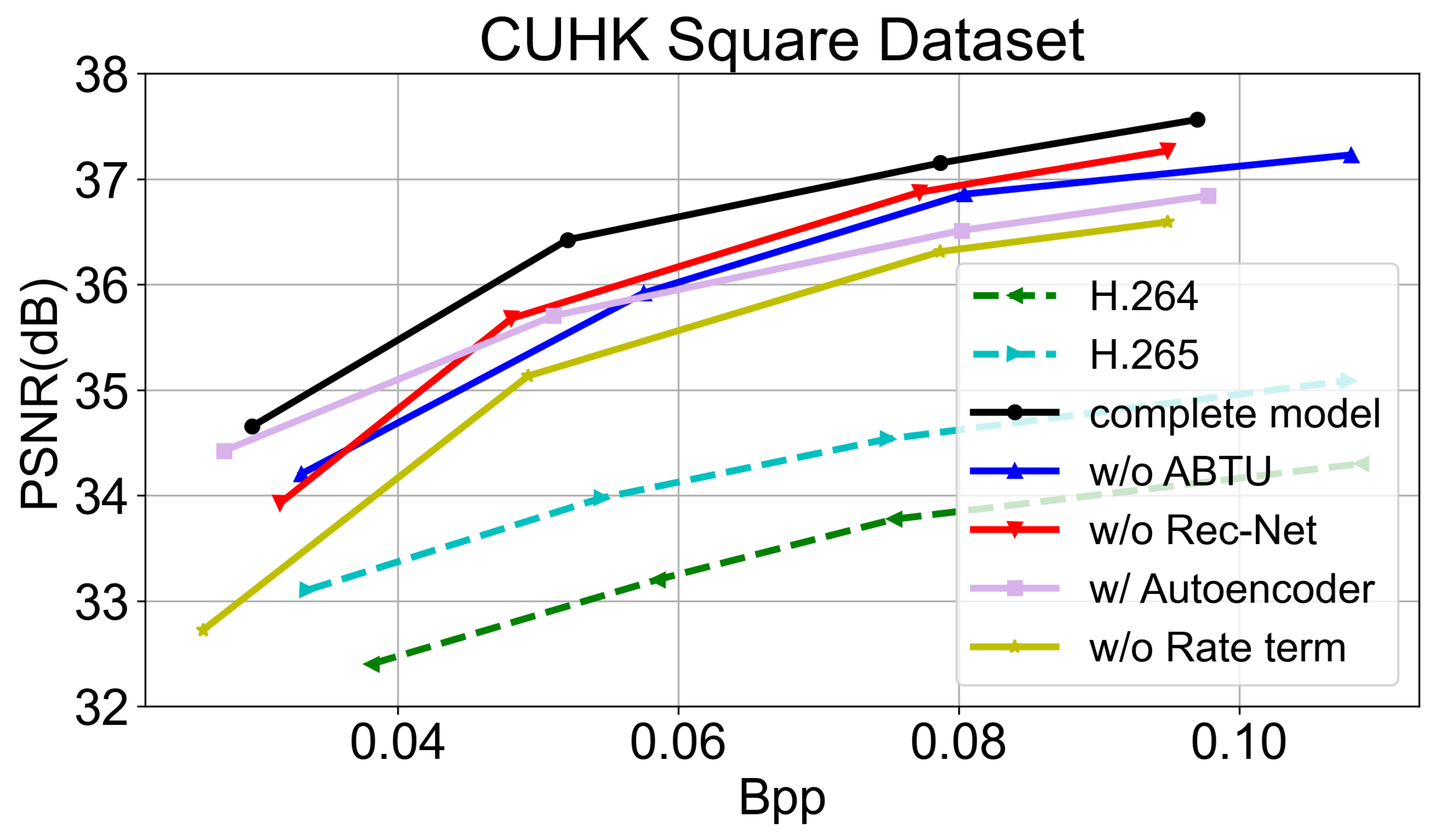}
		\end{center}
		\caption{Illustration of the results of the ablation experiment measured by MS-SSIM and PSNR. ABTU denotes adaptive background template updating.}
		\label{fig:11}
	\end{figure*}

    Based on background fluctuations, the handcraft equal-interval updating is replaced with Adaptive Background Template Updating (\textbf{ABTU}). Unlike other one-step frame decoding methods, we are inspired by \cite{xiong2018learning} to enhance the frame quality through \textbf{Rec-Net}. Besides, the rate term \textbf{$H(v_t, q_t)$} is introduced into Eqn. 5 to optimize rate-distortion simultaneously to further improve the compression ratio. Furthermore, to demonstrate the effectiveness of motion-based residual compression, an \textbf{autoencoder} is applied to directly compress the foreground (denoted as ``w/ autoencoder" in Fig.~\ref{fig:11}), which is achieved as in Fig.~\ref{fig:10} by replacing Step 3-5 in Sec.~\ref{sec:3.1} with the following three steps: (1) The foreground $f_{t}$ is encodes into bitstreams $s_{t}$ directly. (2) The bitstreams $s_{t}$ are quantized as $q_{t}$. (3) $q_{t}$ is decoded into $\overline{f}_{t}$ directly. The specific foreground codec (encoder and decoder) is provided in the appendix.
    
    To demonstrate the importance of the above four components, we have designed the following five models: (1) complete model;  (2) the model without ABTU (update every 50 frames); (3) the model without Rec-Net; (4) the model with autoencoder; (5) the model without rate term $H(v_t, q_t)$. The above five models are trained and tested on the CUHK Square dataset. As shown in Fig.~\ref{fig:11}, the lack of Rec-Net and ABTU downgrades performance to some extent. What's more, we found that the degeneration caused by removing ABTU is pronounced. This is because the information density of surveillance video is very low and the fixed handcrafted updating frequency we set is too high, which reduces the compression ratio required to maintain the same compression quality. Furthermore, although the performance of the autoencoder method is close to the motion-based method at low $bpp$, the gap between their performance widens as $bpp$ increases. The main reason for it is that the autoencoder method processes each frame independently and does not make full use of the temporal information of the foreground. The absence of the rate term may cause a sharp deterioration in performance, which indicates that it is largely insufficient to just optimize the distortion term for compression. In other words, balancing the rate and distortion is very important for compression systems.

    \subsection{Compression Stability Analysis}\label{sec:5.5}
\begin{figure}[h]
        \begin{center}
            \includegraphics[width=0.95\linewidth]{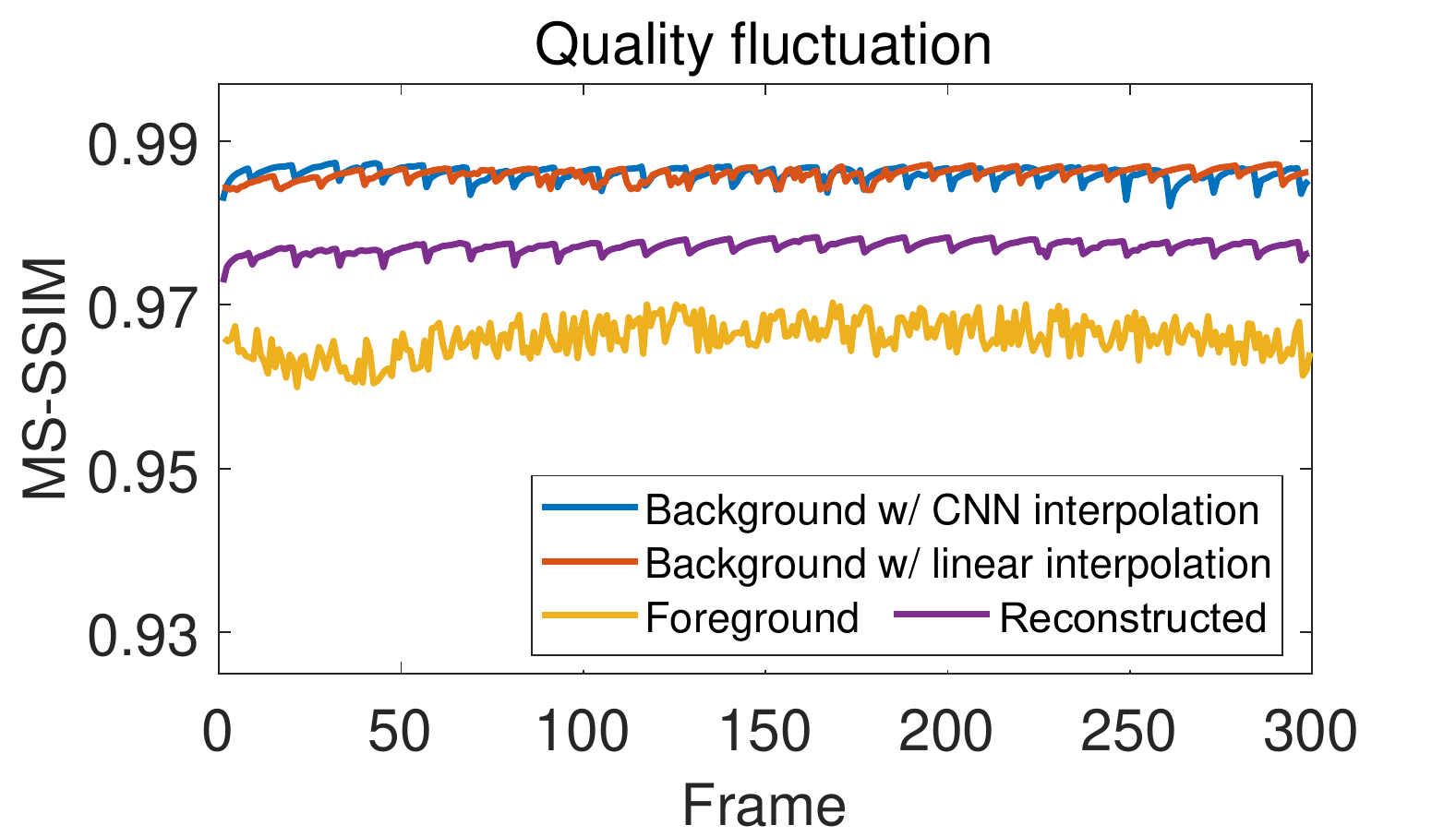}
        \end{center}
        \caption{Illustration of the compression quality fluctuation of background $\overline{b}_{t}$, foreground $\overline{f}_{t}$ and reconstructed frame $\overline{x}_{t}$ within a range of frames.}
        \label{fig:12}
    \end{figure}

\begin{figure}[h]
        \begin{center}
            \includegraphics[width=0.95\linewidth]{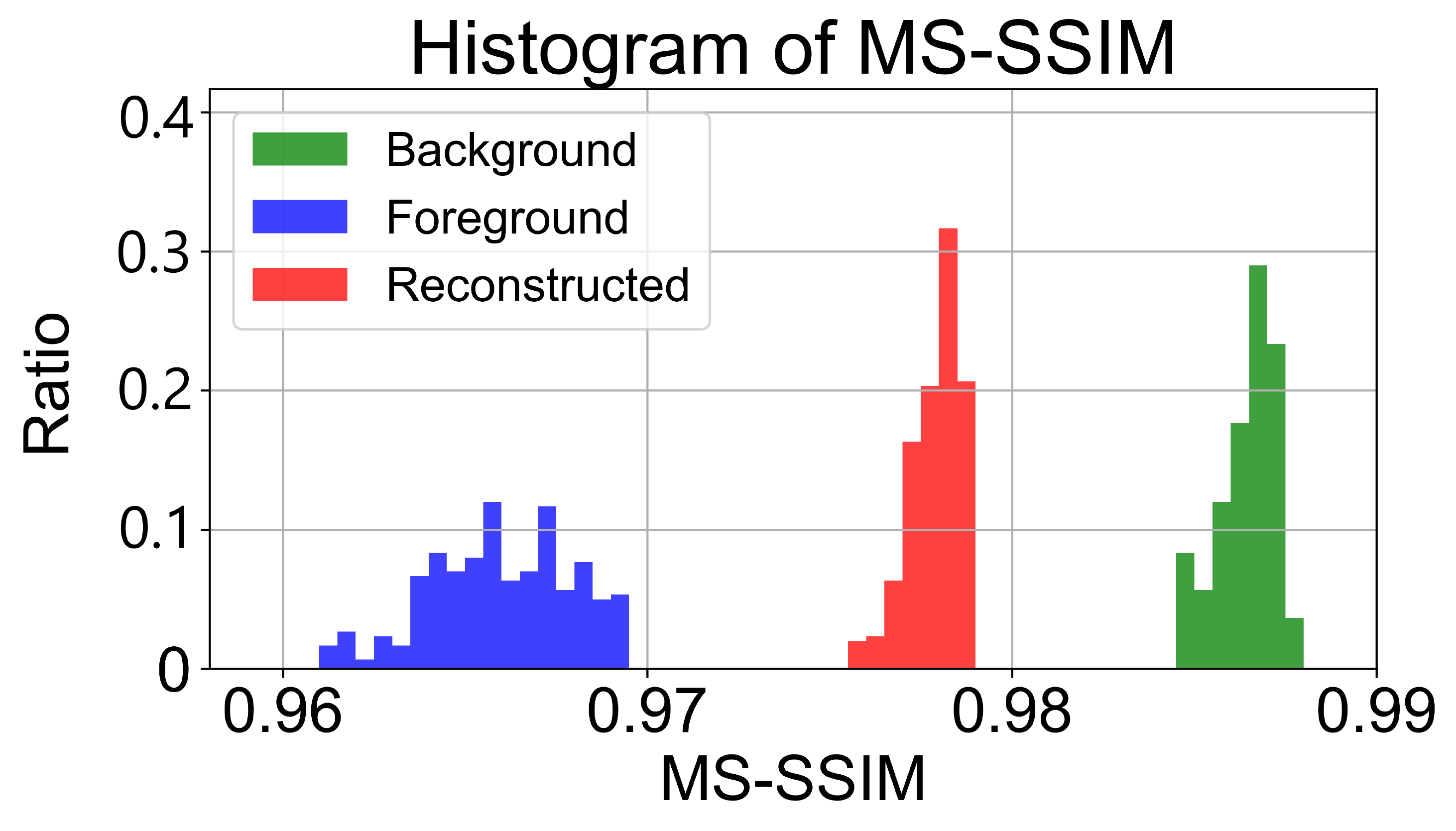}
        \end{center}
        \caption{Illustration of the statistical histogram of MS-SSIM of the background $\overline{b}_{t}$, foreground $\overline{f}_{t}$ and reconstructed frame $\overline{x}_{t}$. Note that the background is implemented with linear interpolation.}
        \label{fig:13}
    \end{figure}

    The proposed compression algorithm is expected to have strong stability, that is, the compression effects of different frames are similar, without large fluctuations. A total of 300 frame are selected to compare background $\overline{b}_{t}$, foreground $\overline{f}_{t}$ and reconstructed frame $\overline{x}_{t}$ with original ones. The background is obtained by linear interpolation described in Eq.~3 and the CNN-based interpolation method proposed in \cite{wu2018video}, respectively. As shown in Fig.~\ref{fig:12}, background $\overline{b}_{t}$ and reconstructed frame $\overline{x}_{t}$ have only small fluctuations, while foreground $\overline{f}_{t}$ has relatively large fluctuations due to the drastic changes that happened in the foreground regions. In addition, we surprisingly found that using the simplest linear interpolation and complex CNN-based interpolation, the stability of the two is very similar, mainly because the following two points: (1) The surveillance video usually has a static background, and the backgrounds at different times are highly similar; (2) The limiting factor of background stability is mainly residual coding, and the interpolation method is necessary but not decisive. Therefore, the linear interpolation is applied by default in our paper. As the statistical histogram of MS-SSIM shows in Fig.~\ref{fig:13}, it can be seen that the compression of our method only fluctuates within a certain range. Besides, the histograms of background and reconstructed frames are pyramidal, which shows relatively strong stability. 
    
    \begin{figure*}[h]
        \begin{center}
            \includegraphics[width=1\linewidth]{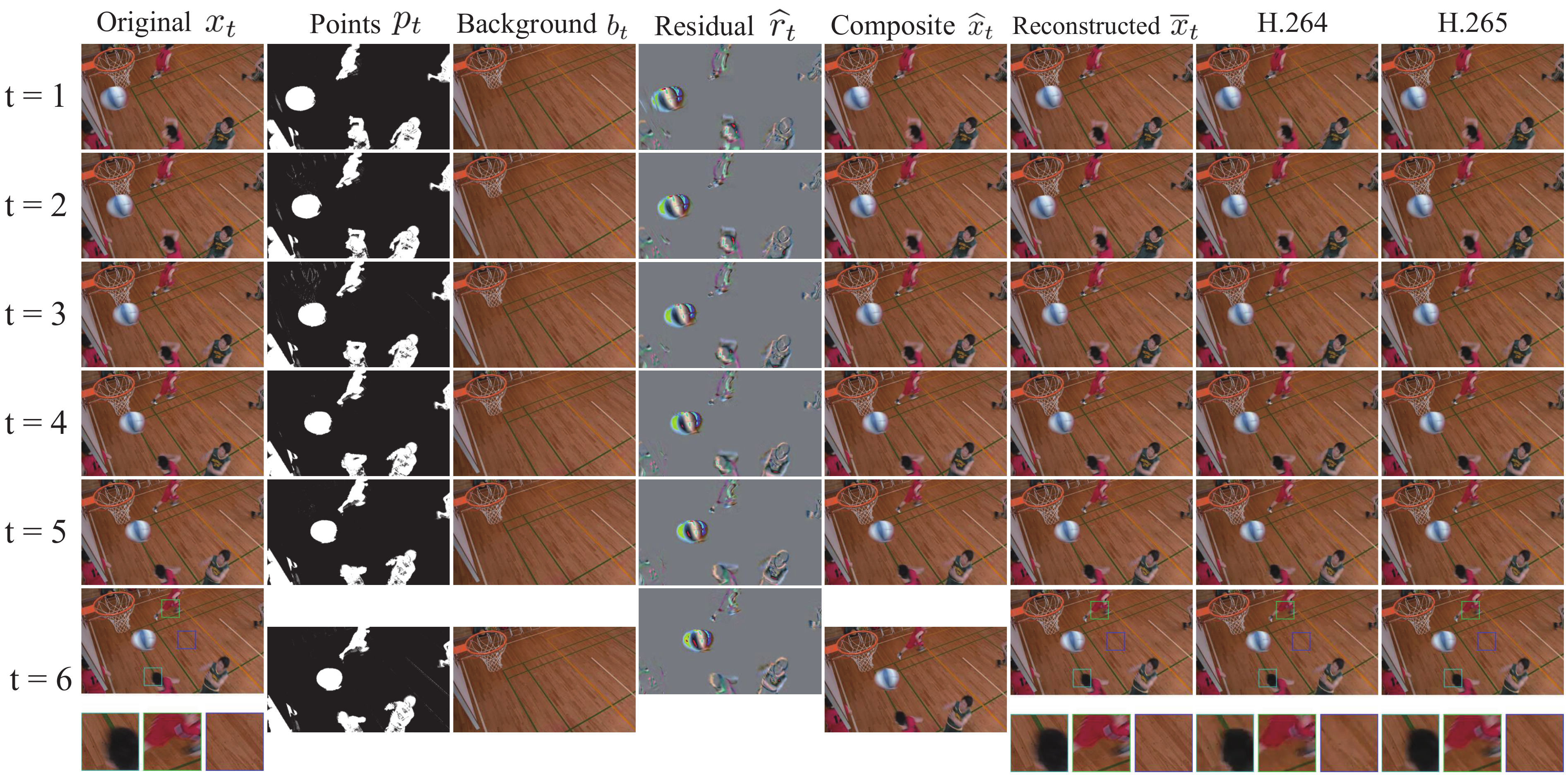}
        \end{center}
        \caption{Illustration of the intermediate results and the compression results of other algorithms. We zoom in on the local details of Frame 6, from which we can see that our compression results are visually better.}
        \label{fig:14}
    \end{figure*}
    
    \subsection{Visual Comparison}
    As shown in Fig.~\ref{fig:14}, we present the intermediate results of compression, including the original frame $x_{t}$, foreground points $p_{t}$, reconstructed background $\overline{b}_{t}$, residual $\widehat{r}_{t}$, composite frame $\widehat{x}_{t}$, and reconstructed frame $\overline{x_{t}}$. Comparing $\overline{b}_{t}$ with the original frame, we found that linear interpolation works well. Visually, the background $\overline{b}_{t}$ is almost identical to the background in the original frame, without severe distortions such as artifacts, blurs, and boundaries.   
    
    In the composite frame $\widehat{x}_{t}$, the composition of the foreground and background template is not very harmonious due to obvious boundaries. However, in the reconstructed frame optimized by Res-Net, the boundaries are completely eliminated, and the transition between foreground and background is very consistent.
    
    Finally, we gave a comparison between our compression results and the commonly used conventional H.264 and H.265. In terms of surveillance video, since these conventional methods are not specifically designed based on the low information density characteristic of surveillance video, their results are not as pleasing as ours visually. In our method, some details of the video, such as the athlete's hair, the outline of the sports pants and the holes in the floor are well preserved.
    
    \linespread{1.2}
\begin{table*}[htbp]
	\centering
	\caption{Sharpness and Preference Analysis}
	\label{table:1}
	\begin{tabular}{l|c|c|c|c|c|c}
		\hline
		\multirow{2}{*}{\ } & \multicolumn{3}{c|}{Sharpness} & \multicolumn{3}{c}{Preference (\%)}\\
		\cline{2-7}
		\quad & 0.046 $bpp$ & 0.087 $bpp$ & 0.132 $bpp$ & 0.046 $bpp$ & 0.087 $bpp$ & 0.132 $bpp$\\
		\hline
		H.264 & 7.78 & 8.09 & 8.95 & 2 & 8 & 12\\
		\hline
		H.265 & 10.34 & 10.96 & 11.60 & 12 & 26 & 26\\
		\hline
		Composite $\widehat{x}$ & 9.89 & 10.32 & 11.03 & 24 & 20 & \textbf{34}\\
		\hline
		Reconstructed $\overline{x}$ & \textbf{11.48} & \textbf{12.47} & \textbf{13.56} & \textbf{62} & \textbf{46} & 28\\
		\hline
	\end{tabular}
\end{table*}
\linespread{1.0}

    \subsection{More Evaluation Metric}
\begin{figure}[h]
        \begin{center}
            \includegraphics[width=1\linewidth]{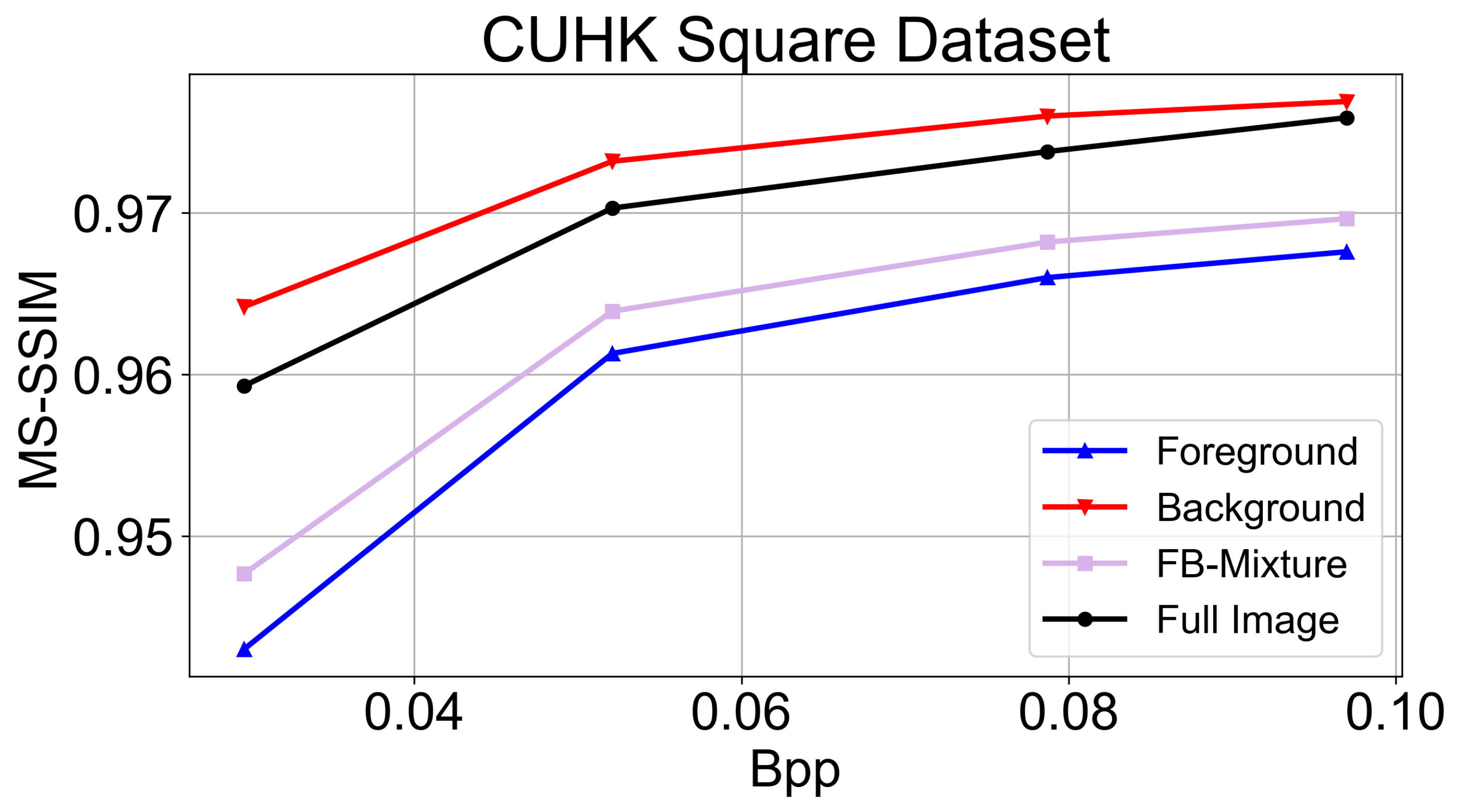}
        \end{center}
        \caption{Illustration of the rate-distortion for foreground, background, full image and new evaluation metric FB-Mixture.}
        \label{fig:15}
    \end{figure}

    The rate-distortion of foreground and background compression is analyzed respectively to demonstrate the idea of parallel compression. As shown in Fig.~\ref{fig:15}, the performance of the background is much better than the foreground, and the performance of the full image is in between but is close to the background. However, the ratio of the foreground in the surveillance video is generally low, and it is not fair to calculate MS-SSIM directly on the full reconstructed image. Therefore, we take into account both the ratio and compression performance, and propose a new quality evaluation metric as shown below to more fairly evaluate the overall compression quality: 
    
    \begin{equation}
    M_{fb\_mixture}=r_b M_f + r_f M_b
    \end{equation}
    
    \noindent where $r_f$ and $r_b$ are the ratio of the foreground and background in the frame, respectively. $M_f$ and $M_b$ are their MS-SSIM. The new evaluation metric reweights contributions of the foreground and background according to their ratios, resulting in a fairer evaluation as the purple line shown in Fig.~\ref{fig:15}.
    
    In the surveillance video, the motion sharpness of foreground objects is a very important evaluation metric. Therefore, we compared our composite frame $\widehat{x}_{t}$ and reconstructed frame $\overline{x}_{t}$ with H.264 and H.265 under three different $bpp$ settings using Laplacian as sharpness metric. As shown in Table.~\ref{table:1}, the performance of H.265 is close and slightly better than composite frame $\widehat{x}_{t}$, but much worse than reconstructed frame $\overline{x}_{t}$, which demonstrates the importance of post-processing. 
    
    Furthermore, a preference study was conducted using Mike CRM platform \cite{mikecrm} to quantitatively evaluate the subjective perceptual quality of our composite and reconstructed results in comparison with H.264 and H.265. Specifically, we considered 3 different $bpp$ settings and invited volunteers to pick up the two with the best subjective perception. A total of 50 valid responses were received from 27 volunteers and the preference ratios were calculated accordingly. As shown in Table.~\ref{table:1}, our method is clearly preferred, especially at low $bpp$.
    
    \subsection{Run efficiency analysis}
    The average time to encode and decode a frame with a resolution of 832 $\times$ 480 is 113.4 ms (8.8 $fps$) and 89.3 ms (11.2 $fps$), running on the GeForce GTX 1080 Ti. The encoding speed of our method is much faster than the most advanced end-to-end DVC \cite{lu2019dvc} (2.22 $fps$) because we only need to compress the foreground regions of each frame. The learning video compression method proposed in \cite{rippel2018learned} was tested on a hardware platform with better performance (V100) and smaller video size (640 $\times$ 480), and their encoding and decoding speeds are 10 $fps$ and 2 $fps$, respectively. Obviously, our run efficiency is significantly higher than theirs, especially the decoding speed. Recently, many neural network acceleration methods have been proposed, which can be applied to our work to improve the encoding and decoding speed, but this is beyond the scope of our paper. 
    
    In Table $\ref{table:2}$, the average time of the separation, compression and composition of foreground and background is presented, and some key steps in foreground compression are also included. The results show that the complexity of foreground compression is much higher than background compression, which is consistent with our intention of focusing more attention on the foreground. In addition, we found that the main computational burden of foreground compression lies in motion compensation, while optical flow computing, motion and residual coding are relatively lightweight operations. More importantly, compared with the block-based method used for the full image, our optical flow computing is only performed on the foreground, which helps to focus limited computing resources on the more important regions. 
    
    \linespread{1.2}
	\begin{table}[htbp]
    	\centering
    	\caption{Complexity Analysis}
    	\label{table:2}
    	\begin{tabular}{l|c}
        	\hline
        	Operation ($Step$) & Running time ($ms/frame$) \\
        	\hline
        	Encoding process & 113.4 \\
        	Decoding process & 89.3 \\
        	\hline
       	 	Foregroun-background seperation ($S1$) & 39.2 \\
        	Background compression ($S2$) & 11.8 \\
        	Foreground compression ($S2-5$) & 78.4 \\
        	Two-stage deconding ($S6$) & 33.8 \\
        	\hline
        	Motion estimation & 22.5 \\
        	Motion compensation & 31.3 \\
        	Information codec & 12.3 \\
        	\hline
    	\end{tabular}
	\end{table}
	\linespread{1.0}

    \section{Discussion}\label{sec:6}
    Conventional video compression algorithms have been widely developed and applied in recent decades. In recent years, many DNN-based compression algorithms have been proposed, but their architecture is basically the same as the conventional architecture, except that the components in the conventional architecture are now replaced with DNNs. The approach we propose in this paper is different from the conventional architectures, which is mainly designed for surveillance video compression. In our architecture, the background is compressed through template updating and background interpolation, which resembles the random access coding structure. The foreground is compressed through motion-based residual compression, which is similar to a low-latency coding scheme.
    
    We firmly believe that our proposed method can inspire researchers to propose new algorithms with wider application fields and superior compression performance. Furthermore, it seems promising to combine high-level vision, such as semantic segmentation, object classification and object detection with image/video compression.

    \section{Conclusion}\label{sec:7}
    In this paper, we have proposed a new video compression method that uses an adaptive Gaussian mixture model to extract the foreground and background of videos separately. Besides, we have found the fluctuations of background and creatively designed an adaptive background updating and interpolation algorithm to adapt it to real-world scene changes. Moreover, we have proposed a motion-based method with residual encoding to compress the foreground. Finally, the coarse-to-fine two-stage module has been adopted to achieve the composition of the foreground and background and the enhancements of composite frames. Experimental results show that our method outperforms other algorithms visually and in terms of objective metric MS-SSIM and PSNR. Incorporating high-level vision, such as semantic segmentation, object classification, and object detection into image/video compression may be a new and promising exploration direction.
    
    \appendices
    \section{Network Structure}    
    There are four convolutional neural networks that need to be introduced in detail in the appendix. They are information encoder and decoder, foreground encoder and decoder, compensation network (Com-Net) and reconstruction network (Rec-Net). The specific network structures are as follows. It is noted that ``conv c64-k3-s2 $\downarrow$'' represents a convolution layer with 64 filters of size 3 $\times$ 3 and a stride of 2. In the same way, ``residual'' and ``transposed\_conv'' represent residual block and transposed convolution, respectively.
    
    \subsection{Information Encoder and Decoder}
\begin{figure}[h]
        \begin{center}
            \includegraphics[width=0.90\linewidth]{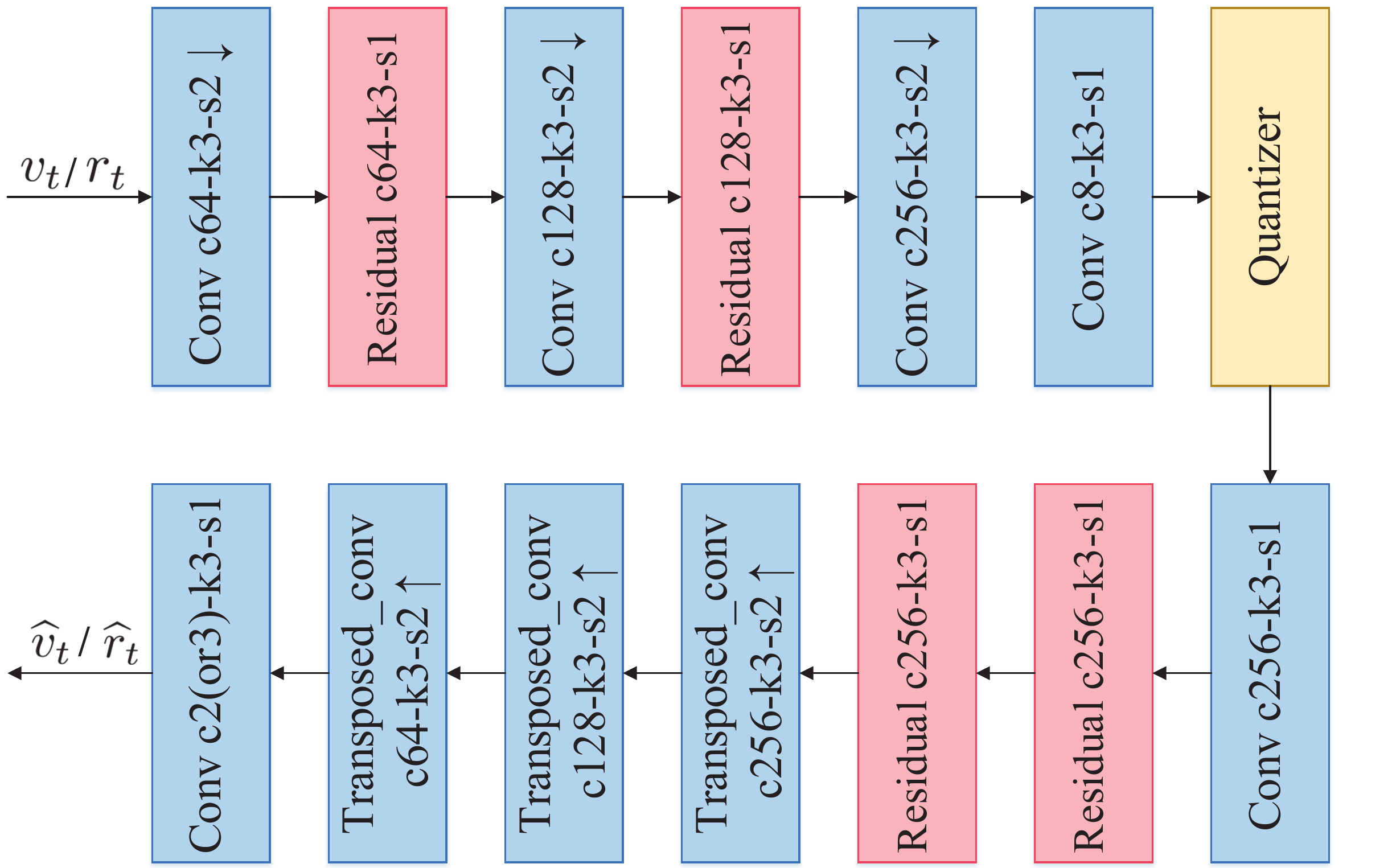}
        \end{center}
        \caption{Illustration of the information encoder and decoder.}
        \label{fig:16}
    \end{figure}

    To increase the compression ratio, the information encoder and decoder mentioned in Step 3 of the Sec.~\ref{sec:3.1} are applied to encode and decode the motion information $v_{t}$ and the residual information $r_{t}$, respectively. The specific network structure is shown in Fig.~\ref{fig:16}.
    
    \subsection{Foreground Encoder and Decoder}
\begin{figure}[h]
        \begin{center}
            \includegraphics[width=1\linewidth]{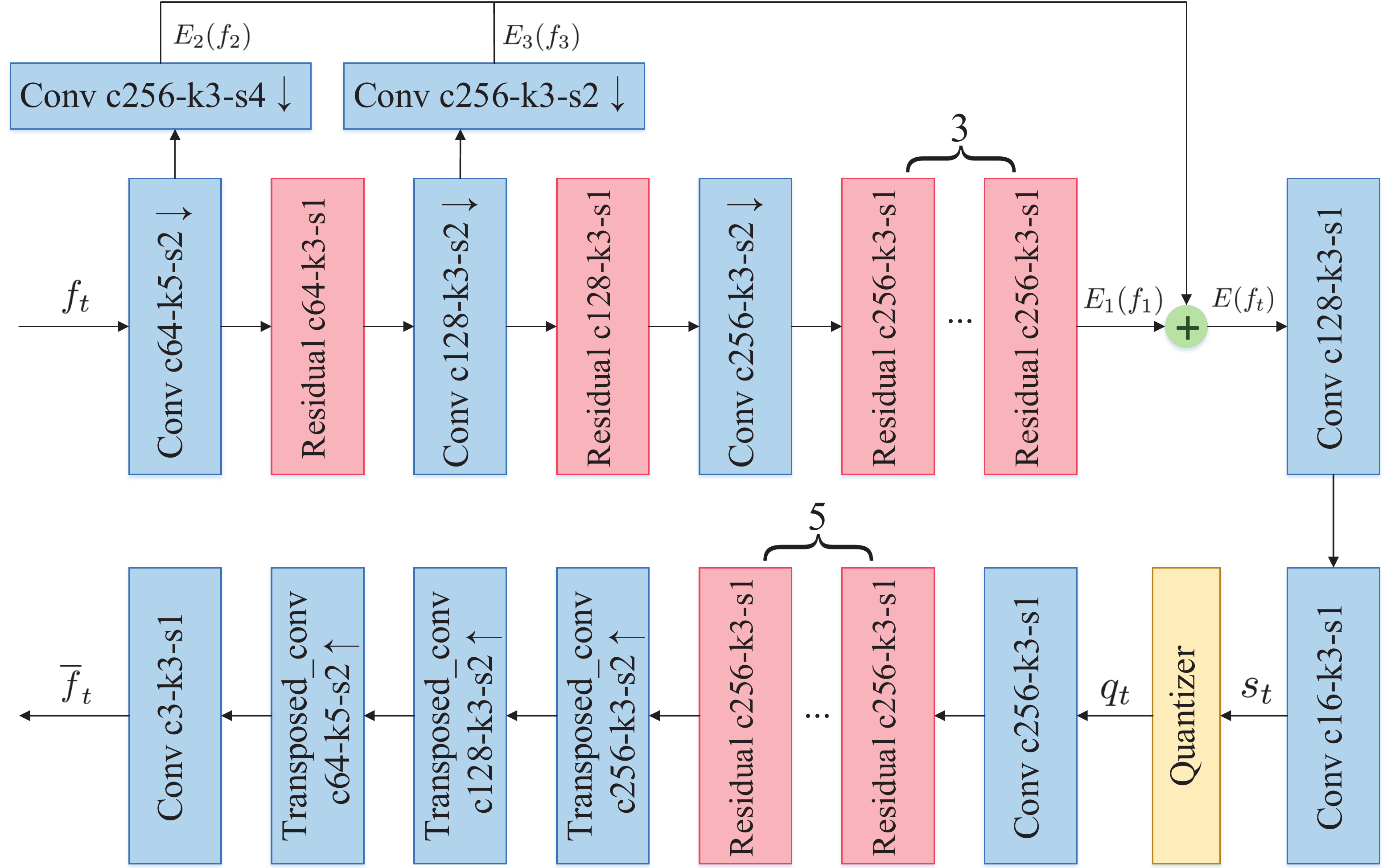}
        \end{center}
        \caption{Illustration of the foreground encoder and decoder.}
        \label{fig:17}
    \end{figure}

    We use the foreground encoder and decoder mentioned in Step 3-5 of Sec.~\ref{sec:5.4} for encoding and decoding the foreground $f_{t}$ directly. As shown in Fig.~\ref{fig:17}, a fully convolutional neural network is used as the encoder, which consists of a crossover stack of several convolutional layers and residual blocks. Furthermore, we adopt the pyramidal decomposition scheme to our encoder. Let $f_m$ denotes the input of the scale $m$ layer, so $f_{1}$ denotes the original foreground $f_{t}$. $E_m(f_m)$ represents the output of the scale $m$. Then we set $m$ to 1, 2 and 3 sequentially and execute encoding individually for each scale. The results of each scale are weighted and summed to produce an output $E(f_t)$ = $E_1(f_1)$ + $\frac{1}{2}E_2(f_2)$ + $\frac{1}{4}E_3(f_3)$. Finally, $E(f_t)$ is convoluted with two convolutional layers to get the output $s_t$.
    
    \subsection{Compensation Network (Com-Net)}
\begin{figure}[h]
        \begin{center}
            \includegraphics[width=1\linewidth]{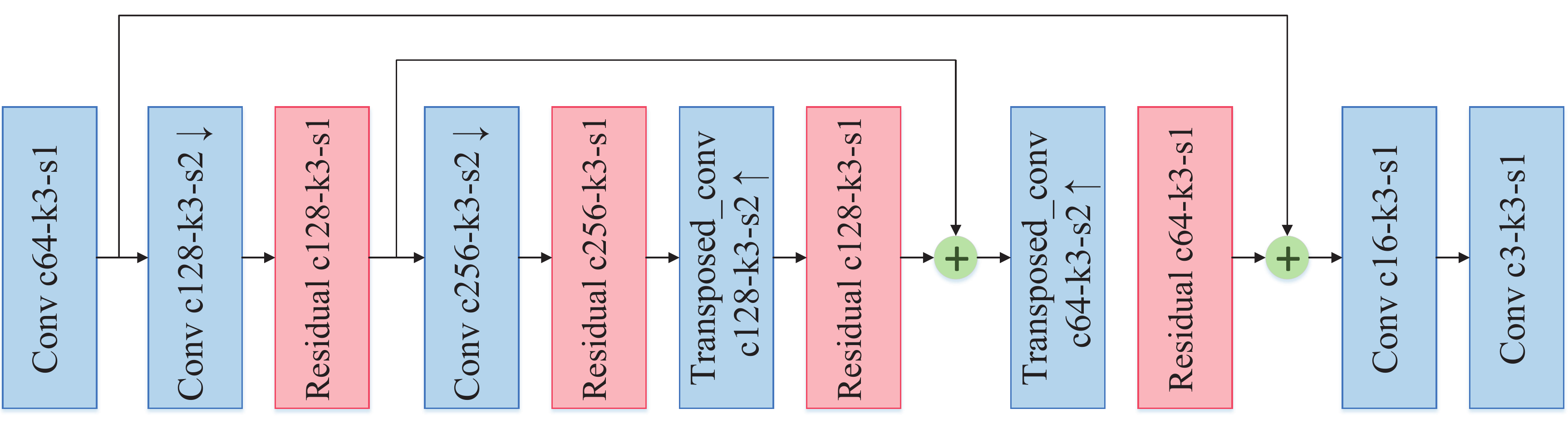}
        \end{center}
        \caption{Illustration of the compensation network.}
        \label{fig:18}
    \end{figure}

    The compensation network used for motion compensation is shown in Fig.~\ref{fig:18}. It inputs the concatenation of the wrapped foreground $w_{t}$, the previous reconstruction foreground $\overline{f}_{t-1}$ and motion information $\widehat{v}_{t}$ to obtain the predicted foreground $\widehat{f}_{t}$.
    
    \subsection{Reconstruction Network (Rec-Net)}
\begin{figure}[h]
    	\begin{center}
    		\includegraphics[width=1\linewidth]{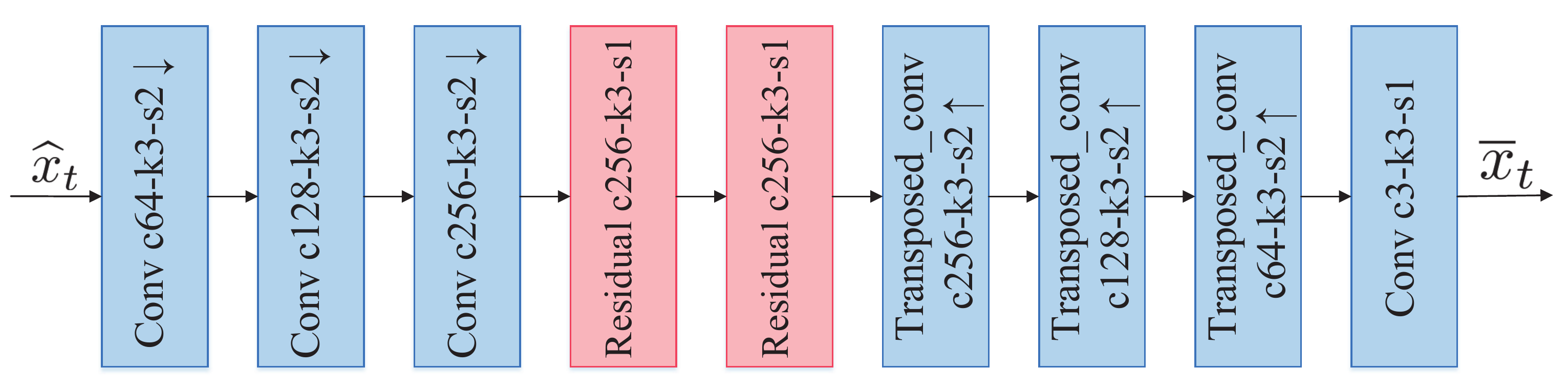}
    	\end{center}
    	\caption{Illustration of the reconstruction network (Rec-Net).}
    	\label{fig:19}
    \end{figure}

    The purpose of Rec-Net is to enhance the quality of the composite frame $\widehat{x}_{t}$ by eliminating block boundaries, blurring and other distortions, resulting in a better visual experience. The specific structure of Rec-Net is shown in Fig.~\ref{fig:19}.
    
    \bibliographystyle{IEEEtran}
    \bibliography{Bibliography-File}

	\clearpage
    \begin{IEEEbiography}[{\includegraphics[width=1in,height=1.25in,clip,keepaspectratio]{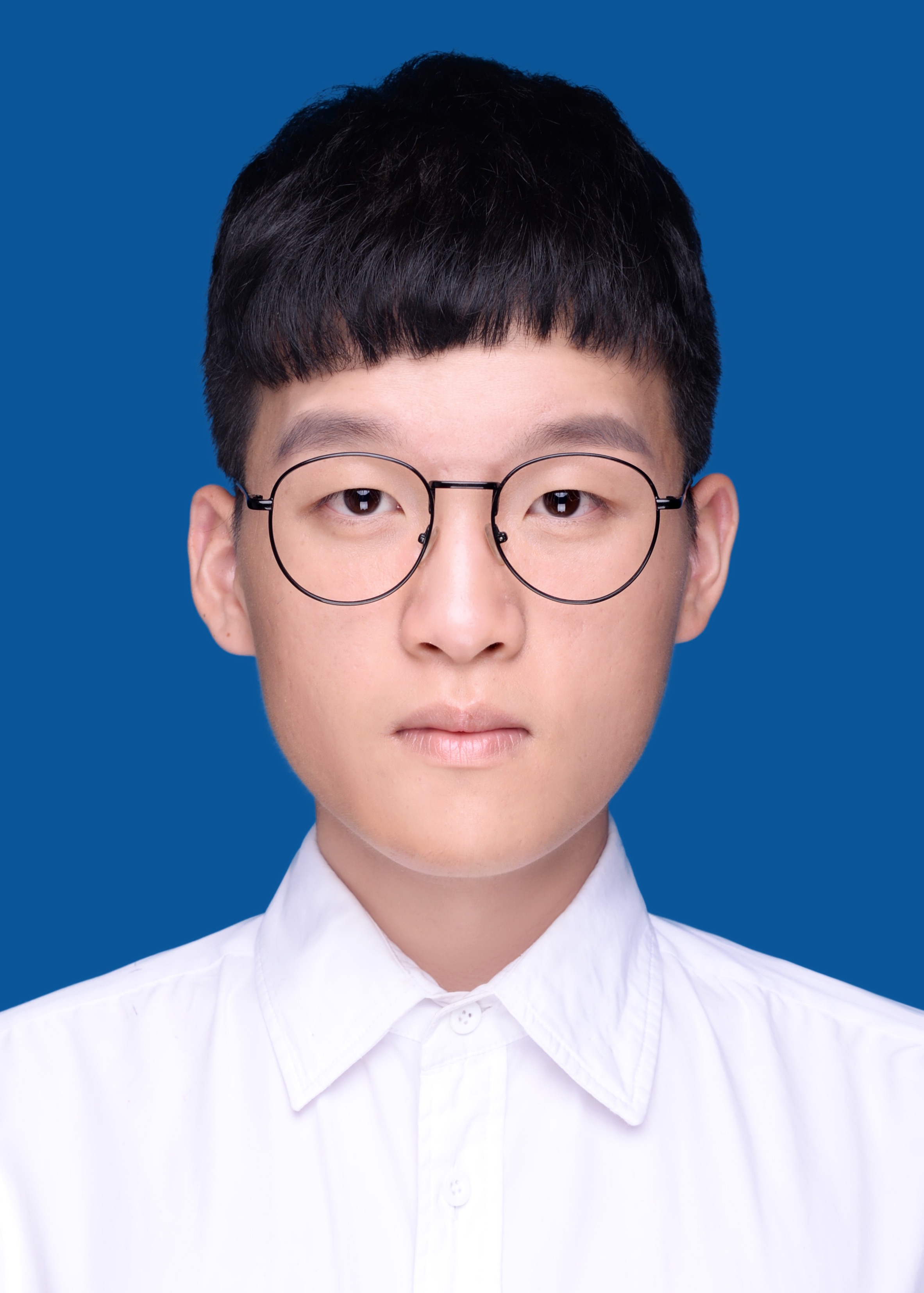}}]{Lirong Wu} is currently pursuing the B.S. degree with the the Department of Information Science \& Electronic Engineering, Zhejiang University, Hangzhou, China. His current research interests include machine learning and image/video processing.
    \end{IEEEbiography}
    
    \begin{IEEEbiography}[{\includegraphics[width=1in,height=1.25in,clip,keepaspectratio]{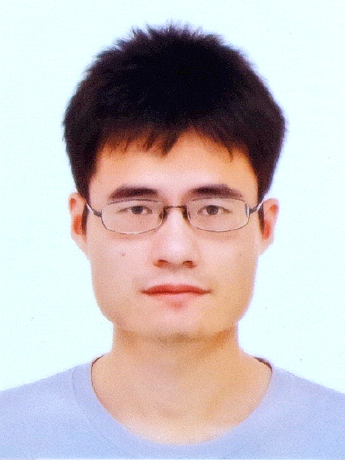}}]{Kejie Huang} (M’13-SM’18) received his Ph.D degree from the Department of Electrical Engineering, National University of Singapore (NUS), Singapore, in 2014. He has been a principal investigator at College of Information Science Electronic Engineering, Zhejiang University (ZJU) since 2016. Prior to joining ZJU, he spent five years in the industry including Samsung and Xilinx, two years in the Data Storage Institute, Agency for Science Technology and Research (A*STAR), and another three years in Singapore University of Technology and Design (SUTD), Singapore. He has authored or coauthored more than 30 scientific papers in international peer-reviewed journals and conference proceedings. He holds four granted international patents, and another eight pending ones. His research interests include architecture and circuit optimization for reconfigurable computing systems and neuromorphic systems, machine learning, and deep learning chip design. He is the Associate Editor of the IEEE TRANSACTIONS ON CIRCUITS AND SYSTEMS-PART II: EXPRESS BRIEFS.
    \end{IEEEbiography}

    \begin{IEEEbiography}[{\includegraphics[width=1in,height=1.25in,clip,keepaspectratio]{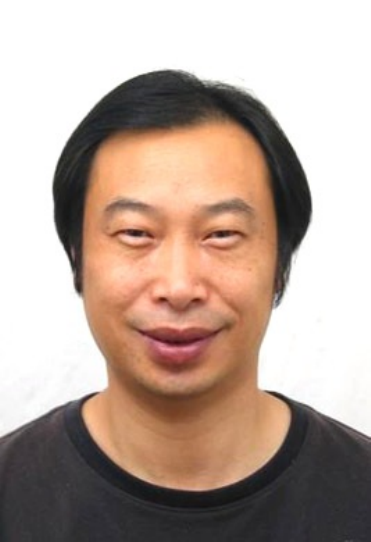}}]{Haibin Shen} is currently a Professor with Zhejiang University, a member of the second level of 151 talents project of Zhejiang Province, and a member of the Key Team of Zhejiang Science and Technology Innovation. His research interests include learning algorithm, processor architecture, and modeling. His research achievement has been used by many major enterprises. He has published more than 100 papers on academic journals, and he has been granted more than 30 patents of invention. He was a recipient of the First Prize of Electronic Information Science and Technology Award from the Chinese Institute of Electronics, and has won a second prize at the provincial level.
    \end{IEEEbiography} 
 
    \begin{IEEEbiography}[{\includegraphics[width=1in,height=1.25in,clip,keepaspectratio]{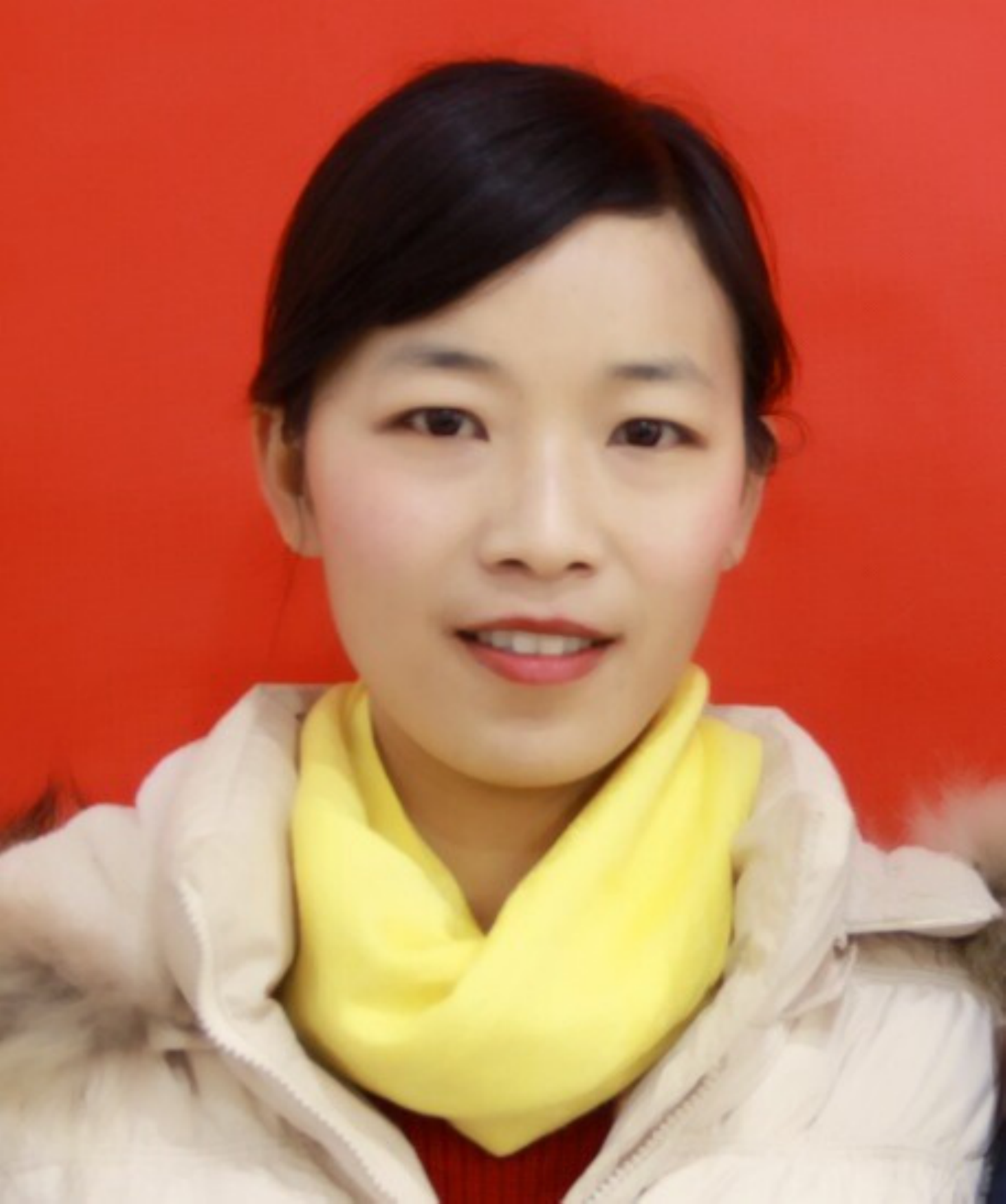}}]{Lianli Gao} is currently a Professor in University of Electronic Science and Technology under the 'UESTC 100 Young Talent' Plan.  She obtained her PhD degree in Information Technology from The University of Queensland, Brisbane, Australia. Her research interests mainly include Deep Learning, Computer Vision, Visual and Language Fusion. She has published 90+ peer-reviewed papers in peer-reviewed conferences and Journals, including TPAMI, IJCV, TIP, AAAI, CVPR, etc. She has been a Guest Editor of the Journal of Visual Communication and Image Representation, and Session Chair of IJCAI'19.
    \end{IEEEbiography}
    
\end{document}